\documentclass[a4paper,11pt]{article}

\usepackage[left=2.4cm,right=2.4cm,top=2.4cm,bottom=2.4cm]{geometry}
\usepackage[T1]{fontenc}
\usepackage[utf8]{inputenc}
\usepackage{tgtermes}
\usepackage{tgheros}
\usepackage{graphicx}
\usepackage{booktabs,tabularx,array,makecell}
\usepackage{amsmath,amssymb}
\usepackage{xcolor}
\usepackage{caption}
\usepackage{enumitem}
\usepackage{microtype}
\usepackage{xurl}
\usepackage{setspace}
\usepackage{placeins}
\usepackage{etoolbox}
\usepackage[colorlinks=true,linkcolor=black,citecolor=blue!55!black,urlcolor=blue!55!black]{hyperref}
\usepackage[super,comma,sort&compress]{natbib}

\hypersetup{
  pdfauthor={Mingkang Liu, Huize Yu, Lei Shen},
  pdftitle={A Density-Matrix Framework for Electronic-Structure Analysis of Electrolytes for Lithium Batteries}
}
\renewcommand{\arraystretch}{1.08}
\setlist{nosep,leftmargin=*}
\pretocmd{\section}{\FloatBarrier}{}{}
\pretocmd{\subsection}{\FloatBarrier}{}{}

\newcommand{\emolstudio}{EMolStudio}
\newcommand{\emolconf}{EMolConf}
\newcommand{\emoles}{EMolES}
\newcommand{\emolesi}{\emoles{}-i}
\newcommand{\emolpipe}{EMolPipe}
\newcommand{\emolagent}{EMolAgent}
\newcommand{\Li}{Li$^+$}
\newcommand{\dHOMO}{$\Delta$HOMO}
\newcommand{\dLUMO}{$\Delta$LUMO}
\newcommand{\BOtot}{\mathrm{BO}^{\mathrm{tot}}_{\mathrm{Li}}}
\newcommand{\BOavg}{\mathrm{BO}^{\mathrm{avg}}_{\mathrm{Li}}}
\newcommand{\PENE}{\mathrm{PENE}}

\title{\bfseries A Density-Matrix Framework for Electronic-Structure Analysis of Electrolytes for Lithium Batteries}

\author{%
\begin{minipage}{0.97\textwidth}
\centering
\normalsize
Mingkang Liu\textsuperscript{1},
Huize Yu\textsuperscript{1}
and Lei Shen\textsuperscript{1,2,3,*}\\[0.7em]
\footnotesize\itshape
\textsuperscript{1}Department of Mechanical Engineering, National University of Singapore, 9 Engineering Drive 1, Singapore 117575, Singapore\\
\textsuperscript{2}National University of Singapore (Chongqing) Research Institute, Chongqing 401120, China\\
\textsuperscript{3}National University of Singapore Guangzhou Research Translation and Innovation Institute, Guangzhou, Guangdong 510700, China\\[0.55em]
\upshape
\textsuperscript{*}Corresponding author. E-mail: shenlei@nus.edu.sg (L.S.)
\end{minipage}%
}
\date{}

\usepackage{longtable,multirow,float,multicol,lastpage}
\newcommand{\emolesI}{\emoles{}-i}
\begin{document}
\pdfbookmark[0]{Main manuscript}{main-part}
\let\savedaddcontentsline\addcontentsline
\renewcommand{\addcontentsline}[3]{}
\begingroup
\setstretch{1.12}
\maketitle

\begin{abstract}
\noindent
Electrolyte reactivity in lithium batteries is shaped by molecular functional groups, Li$^+$ solvation and salt-anion participation. Conventional quantum chemistry is too computationally expensive for systematic analysis of diverse electrolyte molecules and their local solvation environments. Here we present \mbox{\emolstudio{}}, a density-matrix-centered AI platform for electronic-structure prediction and analysis. Its workflow integrates molecular functionalization, explicit Li$^+$ first-shell assembly, density-matrix prediction, and electronic-structure parsing. Applied to 163,655 functionalized molecules and 22,500 first-shell clusters across four lithium salts, we find that 1)~functionalization separates CO$_2$Me, CN, F/CF$_3$, and sulfonyl groups by distinct shifts in frontier levels, electrostatic potential, and Li$^+$--donor contact; 2)~anion identity reshapes frontier-orbital localization, with LiTDI anchoring the highest occupied orbital on the anion across the library. By carrying a unified density-matrix representation from molecular functionalization to salt-resolved solvation shells, \mbox{\emolstudio{}} provides a general platform for understanding and designing battery electrolytes.
\end{abstract}

\vspace{0.6em}

Electrolytes govern ion transport, Li$^+$ solvation and desolvation, and the formation of the solid-electrolyte interphase (SEI) and cathode-electrolyte interphase (CEI) in lithium batteries. The pursuit of higher energy density, faster charging and higher operating voltages places increasing demands on electrolyte chemistry, where molecular stability, ion association, local electric fields and interfacial reactions are closely coupled. Rational electrolyte design therefore requires understanding how molecular structure and salt composition jointly shape the local Li$^+$ solvation environment and its electronic properties\cite{Xu2014Interphases,Meng2022BetterElectrolytes,Hobold2021BeyondCE}.

Functional-group modification is a direct and effective route to improve electrolyte performance. Frontier-orbital energies, electrostatic-potential (ESP) distributions, and Li$^+$ coordination affinity provide a compact electronic-structure readout to rapidly evaluate and screen electrolyte candidates. Fluorinated, oxygen-containing, nitrile, carbonyl, and sulfonyl groups shift these descriptors in chemically distinct ways. In rationally fluorinated ethers such as fluorinated 1,4-dimethoxybutane (FDMB) and fluorinated 1,2-diethoxyethane (F5DEE), cycling performance tracks the number and position of fluorine atoms relative to the coordinating oxygens\cite{Yu2020MolecularSolvent,Yu2022SolventTuning}; surface ESP extrema distinguish solvents into solvating and antisolvent regimes\cite{Wu2023ESP}; and a normalized ion--solvent affinity scale unifies solvation structure, transport, and interphase behavior on a single design axis\cite{Li2025AffinityParadigm}. These molecular electronic characteristics remain constrained by the surrounding coordination environment, as solvation, desolvation, and interfacial electron transfer are collective phenomena of the local Li$^+$ solvation shell rather than isolated molecular properties. The lithium bond and lithium ionic bond formed between Li$^+$ and donor centers give rise to distinct electron-localization and electrostatic signatures within Li-centered clusters, intimately linked to desolvation energetics and interfacial stability\cite{Yao2025LithiumBond}. High-concentration and localized high-concentration electrolytes (LHCEs) exemplify how precisely tuning the first solvation shell and promoting anion participation can effectively redirect electrolyte reactivity and dictate interphasial product composition\cite{Suo2013SolventInSalt,Yamada2014Superconcentrated,Wang2016Superconcentrated,Chen2018LocalizedHCE,Ren2020InnerSolvation,Bergstrom2024IonTransportLHCE}.

Screening these coupled electronic properties across large molecular and multicomponent solvation spaces remains limited by the cost of electronic-structure evaluation. Computational property prediction and force-field molecular dynamics have expanded the accessible candidate, formulation, and liquid-structure spaces. An early virtual-screening study ranked 7,381 carbonate derivatives by scalar reduction descriptors\cite{Halls2010HighThroughput}, and the Electrolyte Genome project automated first-principles ionization potentials, electron affinities, and redox windows across thousands of electrolyte molecules\cite{Qu2015ElectrolyteGenome}. Interpretable regression over elemental compositions identified low solvent oxygen content as a major correlate of lithium-metal cyclability\cite{Kim2023DataDriven}, while a graph-enumerated library of 1,399 ether and carbonyl solvents---adopted here as parent scaffolds---showed that Li$^+$ coordination lowers the solvent LUMO in most cases\cite{Gao2023JACS}. Integrated platforms have further broadened this scope: the Uni-Electrolyte platform couples property prediction, generative design, retrosynthesis planning, and interphase reaction-network analysis over candidate pools exceeding $10^{8}$ molecules\cite{Chen2025UniElectrolyte}, while unified predictive--generative frameworks extend AI-assisted design to complete electrolyte formulations\cite{Yang2026PredictiveGenerativeElectrolyte}. In parallel, classical, ab initio, and machine-learning molecular-dynamics simulations resolve electrolyte solvation structure and transport\cite{Yao2022MDElectrolyte}, and machine-learned force fields now predict density, viscosity, and ionic conductivity for multicomponent formulations\cite{Gong2025Bamboo}. Together, these advances have opened molecular, formulation, and solvation spaces at large scales, but spatially resolved electronic analysis of the resulting candidates still commonly relies on density functional theory (DFT). Combined MD/DFT and cluster quantum-chemistry studies resolve selected systems in detail, showing how Li$^+$ coordination and electrolyte additives modulate solvation motifs, frontier levels, reduction pathways, and interphase-forming tendencies\cite{Hou2019FECSolvation,Chen2020CationAdditive,Hasan2025LiSolvation,Huang2025PolymerSolvation}. Each Li$^+$-centered cluster must nonetheless be assembled as a chemically selected complex or extracted from an MD trajectory and then evaluated by a separate electronic-structure calculation, so exhaustive parent-paired substitution analyses and salt--solvent combinations remain impractical. Equivariant networks for Hamiltonians and related operators have reached high accuracy on molecular and crystalline benchmarks\cite{Yu2023QHNet,Gong2023DeepHE3,Zhouyin2025LEM}, yet chemically diverse electrolyte first shells remain underrepresented in existing training datasets, and the quantities required here---ESP, dipole, bond order, and ELF---depend on the occupied-state density rather than on Hamiltonian eigenvalues alone.

In this work, we present \emolstudio{}, a density-matrix-centered platform for electronic-structure prediction and analysis. Its workflow integrates molecular functionalization, explicit Li$^+$-centered cluster generation, density-matrix prediction, electronic-structure parsing, and spatial visualization. Through these modules, the platform evaluates frontier orbitals, ESP surfaces, orbital localization, Li$^+$--donor electronic contacts (bond order), and the electron localization function (ELF) for both isolated molecules and explicit first-shell clusters. A common density-matrix representation connects density-derived descriptors with reconstructed frontier orbitals. A domain copilot, \emolagent{}, supports the interaction and orchestration of these workflows. We apply \emolstudio{} to two complementary electrolyte design spaces. The first is a library of 163,655 functionalized molecules that maps electronic responses across one to four functionalization steps and eight functional groups. The second is a library of 22,500 explicit Li$^+$ first-shell clusters that examines how LiPF$_6$, LiFSI, LiDFOB, and LiTDI, combined with candidate and background solvents, shape frontier-orbital localization, Li$^+$--donor contacts, and Li-centered electronic-localization features. By connecting molecular functionalization with salt-resolved solvation chemistry through a unified density-matrix representation, \emolstudio{} provides an electronic-structure framework for understanding and designing electrolytes for lithium batteries.

\section{Results}

\subsection{The \emolstudio{} platform}

\emolstudio{} is an electronic-structure analysis workflow tailored to lithium battery electrolytes at library scale, linking molecular design with the electronic structure of \Li{} solvation environments. At the core of the platform lies the density matrix, from which a comprehensive set of descriptors is consistently derived, including frontier orbitals (HOMO and LUMO), ESP, dipole moments, \Li{}--donor bond order, orbital-fragment localization, and \Li{}-centered ELF.

Building upon this density-matrix engine, \emolstudio{} integrates four complementary functional layers (Figure~\ref{fig:platform}). \emolconf{} accepts user instructions, Cartesian coordinates, or simplified molecular-input line-entry system (SMILES) strings and builds the relaxed structures; \emoles{} predicts their density matrices, reaching, after idempotency projection (Methods), density-matrix mean absolute errors of $2.93$--$6.39\times10^{-4}$ and frontier-orbital energy errors within $0.03$~eV on the in-house test sets (Table~\ref{tab:emoles-benchmark}); \emolpipe{} parses each matrix into the descriptors above; and \emolagent{}, a language-model copilot, turns design questions into structured tasks. Methods describes the implementation of each layer.

\begin{figure*}[!htbp]
  \centering
  \includegraphics[width=0.98\textwidth]{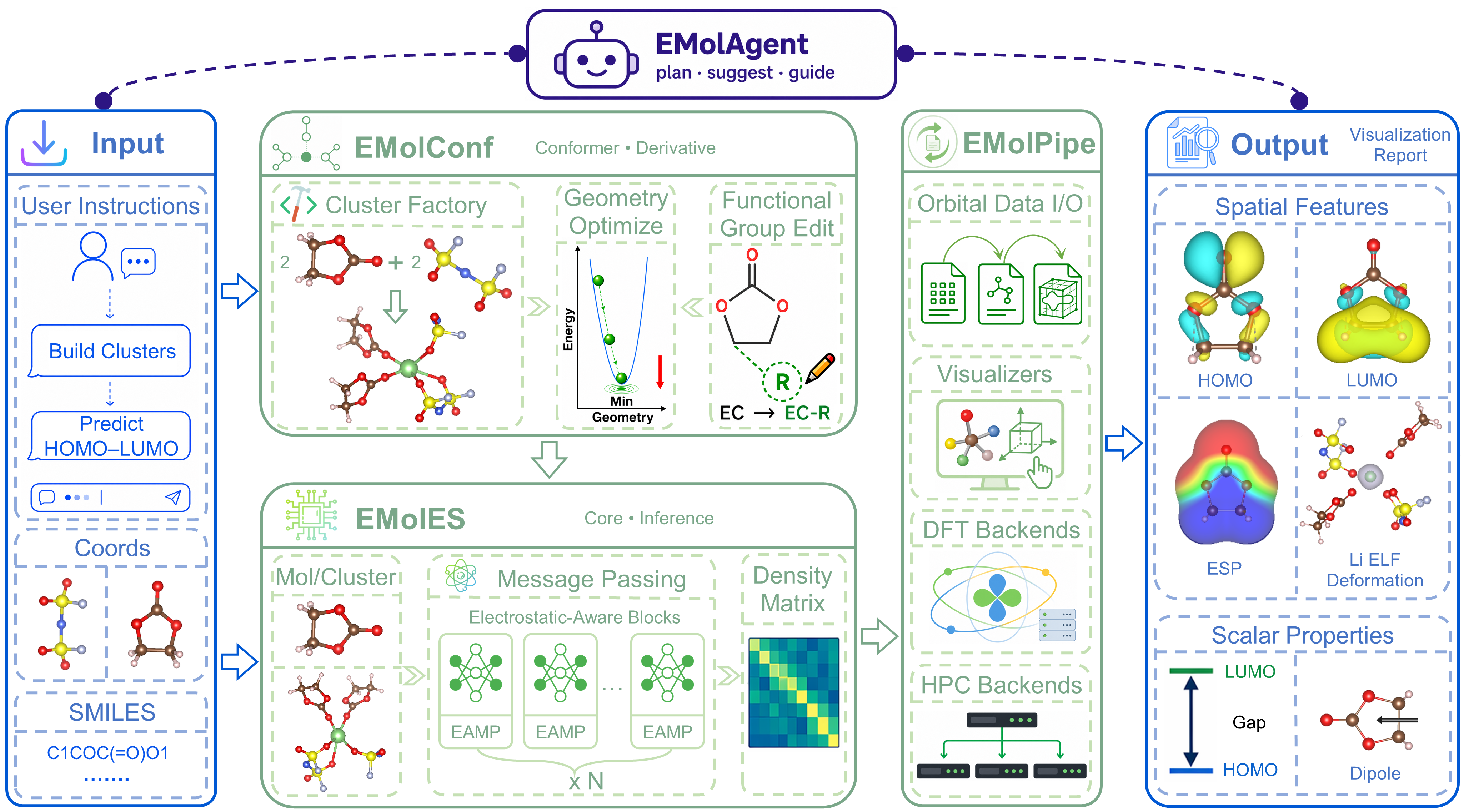}
  \caption{\emolstudio{} workflow for molecular functionalization and solvation electronic-structure analysis. From user instructions, Cartesian coordinates, or SMILES strings, \emolconf{} builds functionalized molecules and explicit \Li{} first-shell clusters and relaxes their geometries. \emoles{} predicts the density matrix of each relaxed structure as a shared representation for downstream analysis, \emolpipe{} parses it into scalar descriptors and spatial visualization objects, and \emolagent{} maps user questions to design tasks and organizes the resulting analysis views. Outputs combine scalar properties with frontier-orbital isosurfaces, ESP-mapped surfaces, and \Li{}-centered ELF deformation descriptors.}
  \label{fig:platform}
\end{figure*}

\subsection{Functionalized-molecule and \Li{} first-shell libraries}

The structure-generation layer, \emolconf{}, turns functional-group edits and \Li{} first-shell compositions into relaxed three-dimensional structures through a shared geometry-preparation and relaxation workflow. Molecular functionalization uses rigid-body fragment attachment to preserve the parent configuration during initial construction, followed by relaxation under a common protocol for parent-matched electronic comparisons. First-shell clusters are assembled chemistry-first, outward from the Lewis-basic sites that may face \Li{}, with steric conflicts resolved before relaxation (Figure~\ref{fig:emolconf}). Each cluster also carries a dielectric constant, which sets the implicit-solvation environment of all downstream cluster calculations. Methods and Supplementary Information Section~S1 give the algorithms, relaxation settings, and quantum-chemical protocols.

\begin{figure*}[!htbp]
  \centering
  \includegraphics[width=0.98\textwidth]{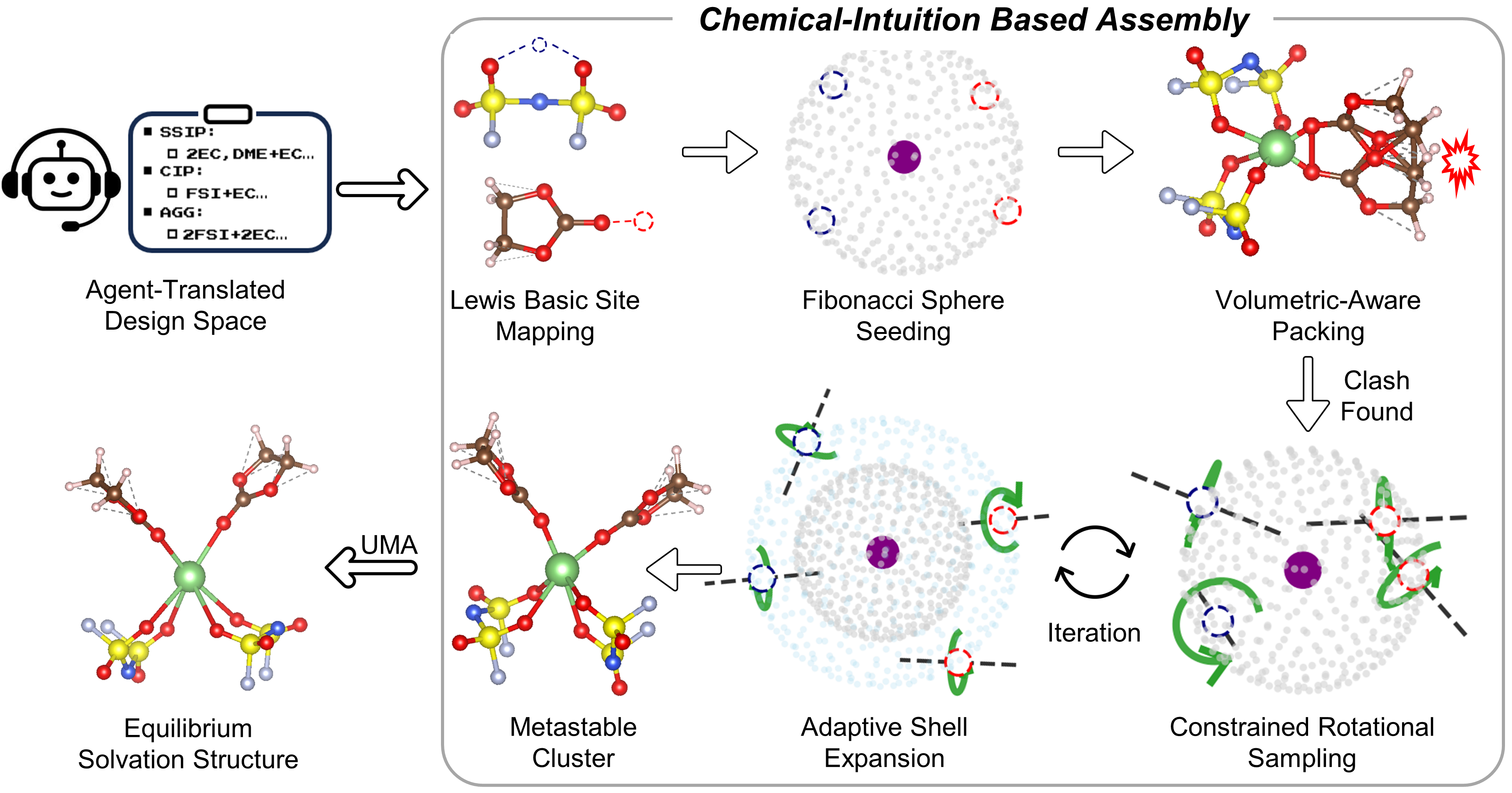}
  \caption{\emolconf{} chemistry-guided assembly of \Li{} first-shell clusters. The Agent-Translated Design Space, the structured task specification produced by \emolagent{} (Methods), specifies the salt environment, solvent composition, and anion-excluded, single-anion-contact, or multianion-contact motif; the examples use ethylene carbonate (EC), 1,2-dimethoxyethane (DME), and bis(fluorosulfonyl)imide (FSI$^-$). The assembly steps shown---Lewis-basic-site mapping, Fibonacci-sphere seeding, volume-aware packing, constrained rotational sampling, and adaptive shell expansion---are defined in Methods; relaxation uses the Universal Models for Atoms (UMA) potential. Atom colors in this and all subsequent atomistic renderings: Li light green, C brown, O red, H white, N blue, S yellow, F pale blue, B pink.}
  \label{fig:emolconf}
\end{figure*}

Using this workflow, we constructed an in-house electrolyte electronic-structure dataset for model development. The molecular subset starts from 1,399 electrolyte-solvent parent molecules\cite{Gao2023JACS} and applies eight functional groups--methyl (Me), cyano (CN), trifluoromethyl (CF$_3$), fluoro (F), methoxy (OMe), methoxycarbonyl (CO$_2$Me), methylsulfonyl (SO$_2$Me), and fluorosulfonyl (SO$_2$F)--giving 4,070 functionalized candidates in each of the neutral and \Li{}-coordinated charge states, of which 4,020 neutral/\Li{} pairs survive geometry optimization and the paired quantum-chemical protocol. The cluster subset contains 7,200 labeled \Li{} first-shell entries (4,608 unique structures) covering 72 common solvents and 10 lithium salts, spanning anion-excluded, single-anion-contact, and multianion-contact coordination. Representative outputs of molecular functionalization and cluster assembly are shown in Figures~S1 and S2 of the Supplementary Information.

\subsection{Density-matrix prediction and its accuracy}

\emoles{} predicts the density matrix used for downstream descriptor evaluation and frontier-orbital reconstruction\cite{Shao2023DensityMatrix,Rackers2023ElectronDensities}. Because that matrix is blocked by atom pairs, the model couples an electrostatic-aware message-passing (EAMP) backbone, which keeps edge features in a bond-aligned SO(2) frame, to an edge-wise atomic cluster expansion (EACE) self-mix that moves high-order equivariant interaction from atoms to atom pairs, and closes with an element-conditioned tensor-product output layer (Figure~\ref{fig:emoles}a--c; Methods). A raw prediction conserves neither electron number nor the closed-shell idempotency constraint, and larger predicted electron-number error (PENE) is associated with larger downstream HOMO/LUMO energy and ESP errors (Supplementary Information, Figure~S3). We therefore apply a projection that restores both constraints; the projected variant, \emolesi{}, is used for the molecular and cluster application analyses reported in this work (Methods).

\begin{figure*}[!htbp]
  \centering
  \includegraphics[width=0.96\textwidth]{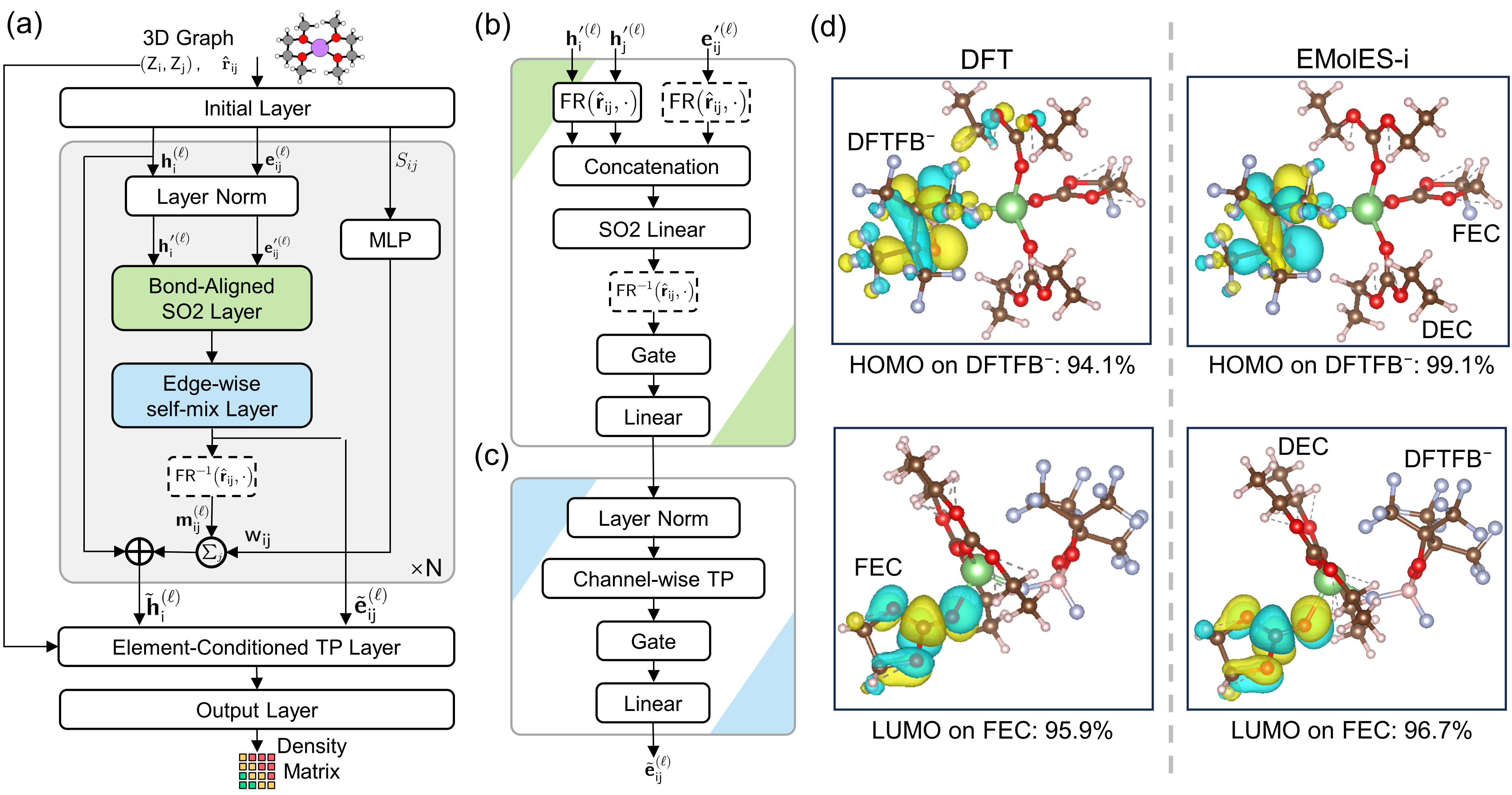}
  \caption{\emoles{} density-matrix architecture and frontier-orbital readout. (a) A three-dimensional atomic graph is encoded by an EAMP backbone with bond-aligned SO(2) message passing, EACE edge-wise self-mix, and an element-conditioned tensor-product layer before density-matrix output. (b) The SO(2) block evaluates edge messages in a bond-local frame. (c) The self-mix block applies channel-wise tensor products to edge features. (d) Comparison between DFT and \emolesi{} HOMO/LUMO isosurfaces for the Li$_1$(DFTFB)$_1$(FEC)$_1$(DEC)$_2$ first-shell cluster, with the dominant fragment contribution reported as orbital-component percentage.}
  \label{fig:emoles}
\end{figure*}

On the in-house molecule and cluster test sets, \emoles{} and \emolesi{} are compared with LEM, SLEM, and QHNet\cite{Zhouyin2025LEM,Yu2023QHNet} using density-matrix MAE and downstream descriptor errors (Table~\ref{tab:emoles-benchmark}; Methods). The unprojected \emoles{} model gives the lowest density-matrix MAE on both test sets and the lowest reported errors among the unprojected models across the cluster-set properties. For density-matrix, frontier-orbital, ESP, and Li-specific metrics, idempotency projection further reduces the errors, and \emolesi{} gives the lowest errors and highest orbital similarities on both test sets. On the recomputed QH9 labels, idempotency projection similarly reduces the density-matrix-derived property errors (Supplementary Information, Table~S4).

\begin{table*}[!htbp]
  \centering
  \footnotesize
  \setlength{\tabcolsep}{3pt}
  \caption{Electronic-structure prediction errors of \emolesi{}, \emoles{}, LEM, SLEM, and QHNet on the in-house molecule and cluster test sets. \emolesi{} denotes the idempotency-projected \emoles{} density matrix; the other rows use unprojected predictions. DM-MAE is the density-matrix mean absolute error; HOMO/LUMO reports frontier-orbital energy MAEs and Sim. the corresponding orbital-coefficient similarities; ESP max/min reports ESP$_{\max}$/ESP$_{\min}$ MAEs; Li def. is the \Li{}-centered ELF deformation-factor MAE and Li BO the average Li-donor bond order MAE. For the molecule set, Li def. and Li BO are averaged over Li-containing structures. Dipole reports the mean absolute error of dipole magnitudes in Debye, with the origin fixed by the stored reference coordinates for all models. Best values are bolded and second-best values are underlined; lower errors and higher similarities indicate better agreement with reference calculations.}
  \label{tab:emoles-benchmark}
  \begin{tabular*}{\textwidth}{@{\extracolsep{\fill}}llrcccccc@{}}
    \toprule
    Set & Model
      & \makecell{DM-MAE\\($10^{-4}$)}
      & \makecell{HOMO/LUMO\\(eV)}
      & \makecell{HOMO/LUMO\\Sim.\ (\%)}
      & \makecell{Dipole\\(D)}
      & \makecell{ESP max/min\\(eV)}
      & \makecell{Li def.\\($10^{-3}$)}
      & \makecell{Li BO\\($10^{-3}$)} \\
    \midrule
    molecule & \emolesi{} & \textbf{6.39} & \textbf{0.027}/\textbf{0.026} & \textbf{99.61}/\textbf{99.62} & \textbf{0.069} & \textbf{0.016}/\textbf{0.015} & \textbf{0.809} & \textbf{0.866} \\
    molecule & \emoles{} & \underline{7.03} & \underline{0.234}/\underline{0.216} & \underline{98.14}/98.23 & 0.241 & \underline{0.048}/\underline{0.035} & 1.337 & \underline{2.382} \\
    molecule & LEM & 7.35 & 0.247/0.237 & 97.50/\underline{98.32} & \underline{0.235} & \underline{0.048}/0.037 & \underline{1.242} & 2.688 \\
    molecule & SLEM & 7.62 & 0.288/0.271 & 97.52/98.18 & 0.268 & 0.050/0.042 & 1.472 & 2.959 \\
    molecule & QHNet & 7.05 & 0.308/0.297 & 97.19/97.07 & 0.314 & 0.059/0.048 & 1.603 & 3.436 \\
    \midrule
    cluster & \emolesi{} & \textbf{2.93} & \textbf{0.030}/\textbf{0.024} & \textbf{98.63}/\textbf{99.81} & \textbf{0.071} & \textbf{0.012}/\textbf{0.011} & \textbf{0.536} & \textbf{0.381} \\
    cluster & \emoles{} & \underline{3.26} & \underline{0.066}/\underline{0.051} & \underline{91.36}/\underline{98.05} & \underline{0.241} & \underline{0.029}/\underline{0.020} & \underline{0.743} & \underline{0.578} \\
    cluster & LEM & 3.62 & 0.087/0.067 & 90.67/97.42 & 0.326 & 0.035/0.025 & 1.059 & 1.018 \\
    cluster & SLEM & 3.60 & 0.071/0.056 & 89.79/97.58 & 0.321 & 0.040/0.025 & 0.761 & 0.700 \\
    cluster & QHNet & 3.48 & 0.169/0.156 & 86.14/92.57 & 0.569 & 0.071/0.052 & 1.514 & 2.161 \\
    \bottomrule
  \end{tabular*}
\end{table*}

As an out-of-training-set case study, we examine a literature-reported first-shell cluster, Li$_1$(DFTFB)$_1$(FEC)$_1$(DEC)$_2$, in which fluoroethylene carbonate (FEC) and diethyl carbonate (DEC) coordinate \Li{} together with the 2,2-difluoro-4,4,5,5-tetrakis(trifluoromethyl)-1,3,2-dioxaborolan-2-uide (DFTFB$^-$) anion\cite{Wang2025LiDFTFB}. This lithium-salt environment is absent from the model training set. The frontier-orbital isosurface topology and fragment-component percentages reconstructed by \emolesi{} agree with the DFT dominant-fragment assignments: the HOMO is mainly distributed on the DFTFB$^-$ fragment, with a contribution of $99.1\%$ against $94.1\%$ from DFT, while the LUMO is mainly distributed on the FEC fragment, with $96.7\%$ against $95.9\%$ (Figure~\ref{fig:emoles}d). This agreement in dominant-fragment assignment supports fragment-resolved orbital analysis for this cluster.

We also follow \Li{} ligand exchange along a two-stage coordination path in which an entering donor joins the first shell while a leaving donor departs (Supplementary Information, Section~S3). \emolesi{} resolves the relative \Li{} binding energy and accompanying Li--O bond-order evolution over the full 1000-frame path, using 30 DFT frames for validation. At those frames, the binding-energy mean absolute error is 0.084~eV ($r=0.995$), compared with 0.380~eV for UMA under the same reference convention. The coupled energy and bond-order profiles place the strongest \Li{} binding near transient coordination-shell closure and provide a continuous view between the DFT validation frames.

\subsection{Electronic effects of molecular functionalization}

Using this workflow, we examine how molecular functionalization changes the electronic properties of an isolated electrolyte solvent before explicit solvation is introduced. This molecular view allows functional-group responses to be considered apart from the salt- and shell-dependent effects examined next. For each functionalized molecule, changes in the HOMO and LUMO energies, ESP extrema, dipole moment, and average \Li{}--donor bond order (BO) are reported relative to the corresponding parent scaffold under a consistent generation, relaxation, and analysis protocol (Methods).

Figure~\ref{fig:smallmol} summarizes these responses. The single-site set contains 19,457 matched structures covering the Me, OMe, F, CF$_3$, CO$_2$Me, CN, SO$_2$Me, and SO$_2$F groups, and the multisite analysis follows one to four functionalized sites across 161,081 matched structures (Supplementary Information, Section~S1.1). Throughout, the frontier levels are read as redox-relevant descriptors rather than as electrochemical windows, which explicit coordination, reaction free energies, interfacial potential, and degradation pathways also shape\cite{Peljo2018HomoLumo,Borodin2013OxidativeStability,Borodin2019Challenges}; likewise ESP$_{\min}$ and ESP$_{\max}$ mark the electron-rich Lewis-basic pocket and the electron-poor region of the molecular surface, which constrain but do not by themselves determine solvation tendency\cite{Wu2023ESP,Li2025AffinityParadigm,Su2019SolvatingPower}.

\begin{figure*}[!htbp]
  \centering
  \includegraphics[width=0.99\textwidth]{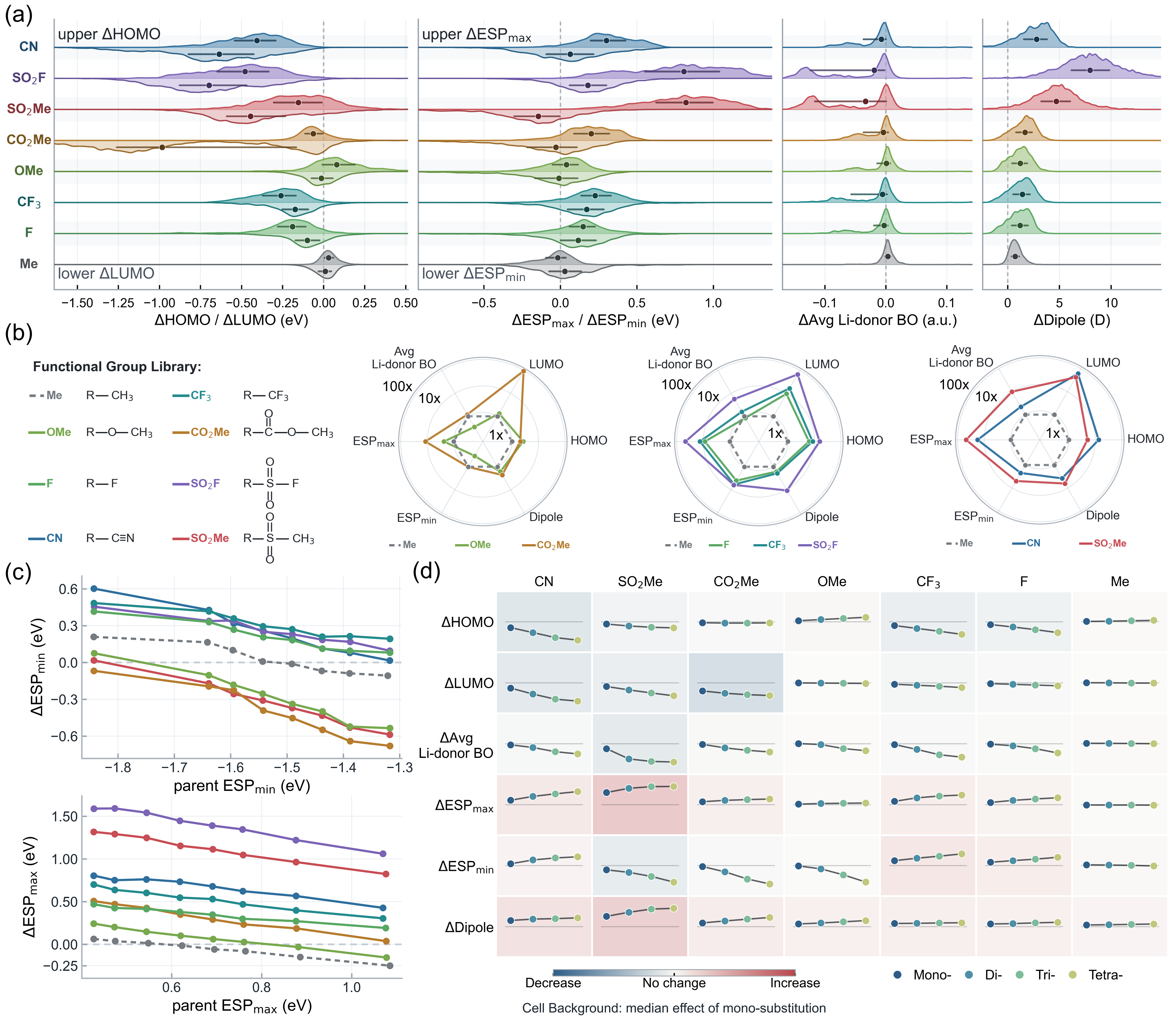}
  \caption{Molecular electronic responses to functionalization. Values are changes relative to the corresponding parent scaffold. (a) Distributions of \dHOMO{}/\dLUMO{}, $\Delta$ESP$_{\max}$/$\Delta$ESP$_{\min}$, $\Delta$\Li{}--donor bond order, and $\Delta$dipole moment for single-site functionalization over 19,457 matched samples; dots and bars mark medians and quantile ranges. (b) Me-referenced radar panels; each axis gives the group median response as a multiple of the corresponding Me median. (c) Parent-conditioned $\Delta$ESP$_{\min}$ and $\Delta$ESP$_{\max}$ trends for all eight groups, colored as in (b). (d) Trajectories from one to four functionalized sites for CN, SO$_2$Me, CO$_2$Me, OMe, CF$_3$, F, and Me; background colors indicate the single-site median direction and points trace the degree of functionalization.}
  \label{fig:smallmol}
\end{figure*}

The eight groups show distinct response patterns after single-site functionalization (Figure~\ref{fig:smallmol}a,b; medians in Table~S5). Me serves as an internal hydrocarbon reference, with near-zero median shifts in the frontier levels and the \Li{}--donor BO, and OMe provides a mild oxygenated reference. Against Me, three broad patterns are apparent: stabilization of the low-lying unoccupied levels by CO$_2$Me, CN, and the sulfonyl groups; HOMO stabilization by SO$_2$F, CN, CF$_3$, and F; and coupled changes in ESP, dipole moment, and \Li{}--donor contact for SO$_2$Me and SO$_2$F.

CO$_2$Me and CN exemplify two distinct electron-accepting motifs. CO$_2$Me shows a response consistent with carbonyl $\pi^\ast$ acceptor character and the largest single-site LUMO stabilization among the eight groups (median \dLUMO{} $=-0.983$ eV), while its \Li{}--donor contact changes more moderately. CN combines the inductive withdrawal of the linear C$\equiv$N unit with its large bond dipole, lowering both frontier levels (median \dHOMO{}/\dLUMO{} $=-0.406/-0.636$ eV), reinforcing the local positive field ($\Delta$ESP$_{\max}=+0.302$ eV), and raising the dipole moment by $+2.81$ D. These responses set a reference for the LUMO localization seen in explicit first shells below\cite{Chen2021IonSolvent,Gao2023JACS}.

CF$_3$ and F display the inductive stabilization and local-field tuning characteristic of fluorination: both lower the HOMO appreciably and raise ESP$_{\max}$ and ESP$_{\min}$, with CF$_3$ giving the larger local-field and \Li{}--donor-contact changes. The scaffold-matched Me/F/CF$_3$ series resolves this graded trend (Supplementary Information, Figure~S7), which tracks the balance among fluorination degree, donor strength, and coordination ability reported for fluorinated and antisolvent formulations\cite{Zhao2022FluorinatedEther,Wu2022Antisolvents}.

The sulfonyl groups show the largest combined changes in local electric field, dipole moment, and \Li{}--donor contact. SO$_2$Me gives the largest median $\Delta$ESP$_{\max}$ in the single-site set ($+0.821$ eV) together with the largest attenuation of the \Li{}--donor BO ($-0.033$), consistent with perturbation of the native \Li{}--donor environment alongside the increased polarity. SO$_2$F responds comparably in ESP$_{\max}$ ($+0.808$ eV) but adds deeper HOMO stabilization and the largest single-site dipole increase ($+7.96$ D). Both are consistent with the reported coupling of polar sulfone media to inner-shell structure and interphase chemistry\cite{Xu2002SulfoneElectrolytes,Ren2018SulfoneLHCE,Chen2018LocalizedHCE,Rahman2023SulfoneDelocalization}.

The strength of these responses also depends on the electrostatic background of the parent scaffold (Figure~\ref{fig:smallmol}c; Table~S6). SO$_2$Me, SO$_2$F, CN, and CF$_3$ amplify the local positive field most effectively on parents whose initial ESP$_{\max}$ is low, and the increment converges once the scaffold is already electron-poor; $\Delta$ESP$_{\min}$ shifts analogously with the initial depth of the parent Lewis-basic pocket. Such scaffold--substituent coupling is consistent with classical Hammett-type relationships and absolutely localized molecular orbital (ALMO) analyses, in which inductive effects, resonance, permanent electrostatics, and fragment-orbital mixing jointly set the frontier levels\cite{Hansch1991SubstituentConstants,Mao2018SubstituentOrbitals,Wilbraham2020OptoelectronicPropertySpace}.

Following each group from one to four functionalized sites places these single-site effects on continuous trajectories (Figure~\ref{fig:smallmol}d, which shows seven of the eight; endpoints in Table~S5, distributions in Figures~S5 and S6). CO$_2$Me deepens its LUMO stabilization from $-0.983$ to $-1.537$ eV while retaining a $\pi^\ast$-dominated response, CN reaches one of the largest cumulative LUMO stabilizations in the set ($-2.482$ eV at four sites) alongside progressively deeper HOMO stabilization, and CF$_3$ and F build up inductive stabilization with increasing fluorination. SO$_2$Me couples the ESP extrema, dipole moment, and \Li{}--donor contact ever more strongly, its median BO response growing from $-0.033$ to $-0.129$, while Me stays flat throughout. For most groups the single-site direction is retained at higher degrees of functionalization but the increments attenuate, so single-site statistics rank the groups and the trajectories show how that response accumulates, often with diminishing increments, within a scaffold.

\subsection{Salt-resolved \Li{} first-shell chemistry}

Molecular functionalization tunes the electronic properties of a candidate solvent, but its reactivity in an electrolyte also depends on the local environment formed around \Li{} by solvent donors, salt anions, and diluents\cite{Yao2025LithiumBond,Fadel2019SolventAnionChargeTransfer}. Explicit first-shell clusters connect the two levels: on the same relaxed structure they resolve which fragment hosts the frontier orbitals, how strongly \Li{} maintains electronic contact with its donors, and how the electronic environment around \Li{} changes. The cluster library comprises 22,500 structures spanning LiPF$_6$, LiFSI, lithium difluoro(oxalato)borate (LiDFOB), and lithium 4,5-dicyano-2-(trifluoromethyl)imidazolide (LiTDI), 5,625 per salt, in three fixed motifs: solvent-rich M$_2$X (two solvents, one anion), diluent-containing MBX (solvent, diluent, anion), and anion-rich MX$_2$ (one solvent, two anions), as detailed in Supplementary Information Section~S1.2, with full statistics in Tables~S7--S10.

\begin{figure*}[!htbp]
  \centering
  \includegraphics[width=0.99\textwidth]{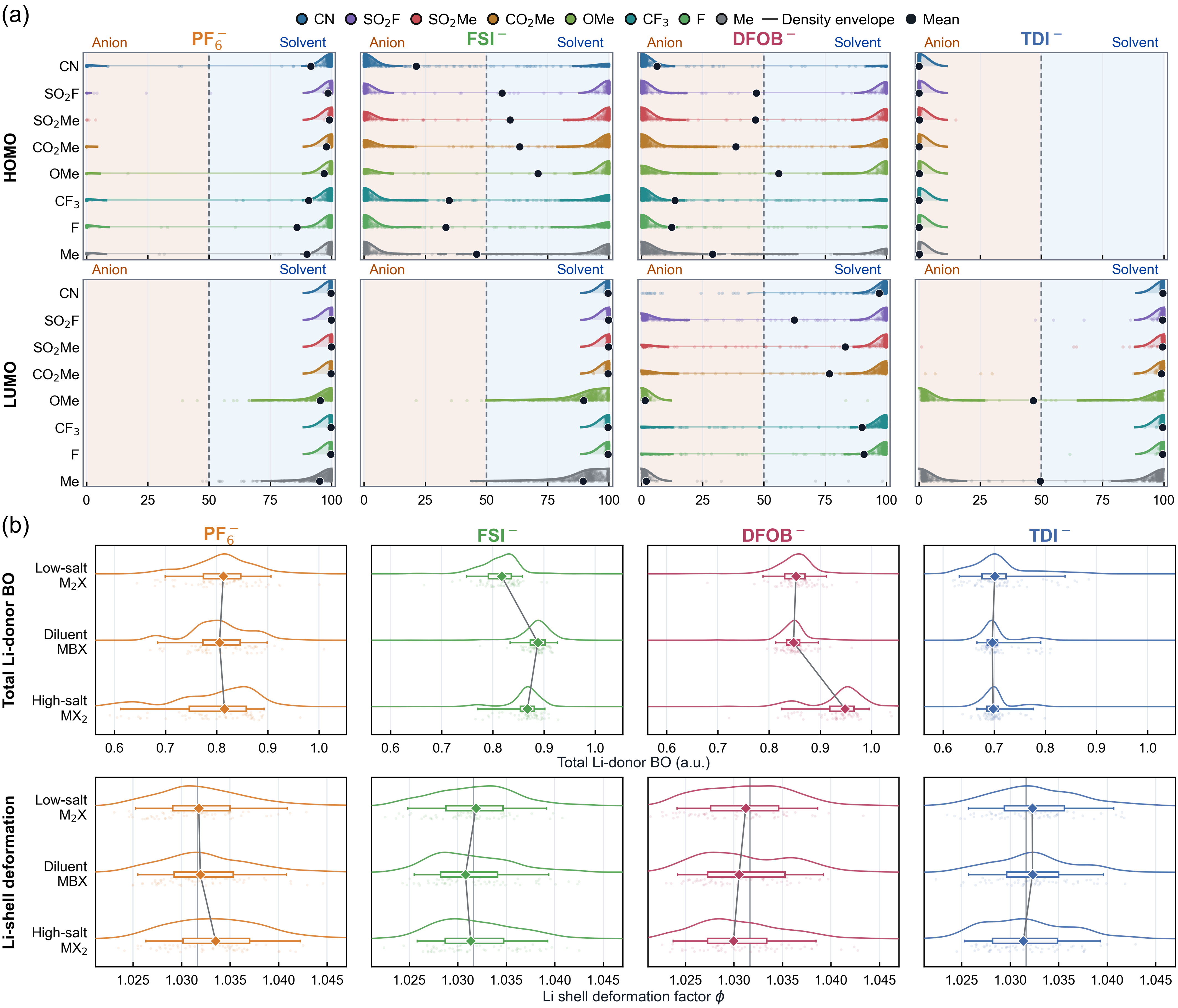}
  \caption{Salt-resolved frontier-orbital hosting and \Li{}-centered descriptors for the complete 22,500-cluster set. (a) HOMO and LUMO solvent-orbital share, defined as $w_{\mathrm{solvent}}/(w_{\mathrm{anion}}+w_{\mathrm{solvent}})\times100\%$, for eight functional groups and four salts. The 0\% and 100\% limits denote anion- and solvent-dominated localization, respectively; points and density envelopes show the per-structure distributions and black dots their means; the dashed line at 50\% is the solvent/anion assignment boundary used for the hosting fractions quoted in the text. (b) Distributions of total \Li{}--donor bond order and \Li{}-shell ELF deformation factor $\phi$ for solvent-rich M$_2$X, diluent-containing MBX, and anion-rich MX$_2$ motifs; diamonds mark medians and the vertical reference in the $\phi$ panels marks the pooled median.}
  \label{fig:cluster}
\end{figure*}

Frontier-orbital hosting differs markedly among the four salts (Figure~\ref{fig:cluster}a; Table~S7). LiPF$_6$ clusters are dominated by solvent-derived frontier orbitals, with solvent-hosted fractions of 94.1\% for the HOMO and 99.9\% for the LUMO. LiTDI shows a contrasting occupied-state pattern: every one of its 5,625 HOMOs is anion-hosted, whereas 91.6\% of its LUMOs remain solvent-hosted. LiFSI sits close to the HOMO partition boundary, with 46.6\% solvent-hosted and 53.4\% anion-hosted structures alongside an almost completely solvent-hosted LUMO (99.9\%), and LiDFOB pairs a predominantly anion-hosted HOMO (71.9\%) with a predominantly solvent-hosted LUMO (71.1\%). These fractions describe the vertical frontier-orbital composition of the intact closed-shell cluster: HOMO hosting locates oxidation-relevant occupied states and LUMO hosting locates reduction-relevant unoccupied states; they do not by themselves set the reduction sequence, which for FSI$^-$ in particular reflects the relaxed reduced state and the dissociative reaction free energy rather than a low-lying vertical LUMO\cite{Suo2013SolventInSalt,Yamada2014Superconcentrated,Chen2018LocalizedHCE,Ren2020InnerSolvation,Peljo2018HomoLumo,Borodin2013OxidativeStability,Borodin2019Challenges}. The LiTDI pattern agrees with recent work on self-compartmented LiTDI electrolytes, in which TDI$^-$ enriches at the cathode interface and raises the HOMO energy of anion-rich solvation structures\cite{Ou2025SelfCompartmented}.

Within the salts of mixed character, frontier-orbital localization also varies with molecular functionalization (Table~S10). For LiFSI, the solvent-hosted HOMO fraction ranges from 20.9\% for CN to 71.3\% for OMe, while the LUMO remains almost entirely solvent-hosted for all eight groups. For LiDFOB, LUMO hosting is strongly group-dependent: 97.9\% of CN structures are solvent-hosted, compared with 0.4\% for OMe and 0.9\% for Me. LiTDI keeps an anion-hosted HOMO for every group, with Me and OMe LUMOs close to the solvent/anion boundary. Salt identity therefore largely sets the dominant localization regime, and within the mixed-character salts functionalization selects the hosting fragment, an electronic-structure counterpart to the reported roles of DFOB$^-$- and TDI$^-$-derived species in interphase construction and anion-derived passivation\cite{Chen2007LiDFOBSalt,Chen2006LiDFOBAdditive,Cha2017LiDFOBMechanism,Paillet2015LiTDIStability,Lindgren2016TDISEI,Li2021AnionDerivedSEI,Wu2022AnionRich,Ruan2023SolventAnionChemistry}.

Total \Li{}--donor bond order condenses the aggregate electronic contact of the first shell into a single quantity (Figure~\ref{fig:cluster}b; Tables~S7 and S8). The salt-pooled medians rank LiFSI (0.863) and LiDFOB (0.858) above LiPF$_6$ (0.811), with LiTDI lower (0.698). Coordination topology has its largest effect for LiDFOB, whose median rises from 0.852 in M$_2$X and 0.848 in MBX to 0.948 in anion-rich MX$_2$; LiFSI peaks in the diluent-containing MBX motif (0.888), whereas LiTDI changes little across motifs (0.696--0.701). Aggregate contact therefore varies primarily with salt identity and topology, consistent with the donor capacity of the anions themselves: the chelating oxalato oxygens of DFOB$^-$ add \Li{} contacts as further anions enter the shell, whereas the charge-delocalized TDI$^-$ anion sustains the weakest aggregate contact.

The \Li{}-centered ELF deformation factor $\phi$ varies over a narrower range: the salt-pooled medians span 1.0306--1.0323, and within each salt $\phi$ shows at most a weak relation to the total bond order ($|r|\leq0.19$, under 4\% of the variance). The two descriptors thus probe largely complementary aspects of first-shell chemistry, the integrated strength of \Li{}--environment contact and the reorganization of electron localization around \Li{}. Functional-group effects on the total bond order stay smaller than the salt and topology contrasts: SO$_2$F gives the lowest pooled median (0.806) and the combined sulfonyl subset (0.814) sits below the remaining groups (0.832) (Table~S9). Across this library, frontier-orbital placement varies more clearly with functionalization, whereas \Li{}-centered contact is more strongly associated with salt identity and local topology.

\section{Discussion}

This work presents \emolstudio{}, a density-matrix-centered workflow integrating molecular functionalization, explicit \Li{} first-shell construction, model inference, and chemical interpretation. Among the unprojected models, \emoles{} gives the lowest density-matrix errors on the in-house molecule and cluster tests, and projection further improves the density matrix and derived descriptors. The workflow is also evaluated on a first-shell cluster containing a lithium-salt environment absent from training and along a 1000-frame \Li{} ligand-exchange path; against 30 DFT validation frames, the binding-energy mean absolute error is 0.084~eV.

Analysis of 163,655 functionalized molecules characterizes the electronic responses associated with eight functional groups. CO$_2$Me provides the largest single-site LUMO stabilization ($-0.983$~eV), consistent with a carbonyl $\pi^\ast$ acceptor response, whereas CN lowers both frontier levels while strengthening the local positive field. F and CF$_3$ show a graded fluorination trend in frontier-level and electrostatic tuning, and SO$_2$Me/SO$_2$F produce the largest combined changes in polarity, ESP, and \Li{}--donor contact. Across one to four functionalized sites, the single-site response direction is generally retained, with diminishing increments at higher degrees of functionalization.

The 22,500 explicit first-shell clusters show how salt identity and coordination topology modulate these molecular trends. LiPF$_6$ retains predominantly solvent-hosted frontier orbitals, whereas all 5,625 LiTDI HOMOs are anion-hosted. LiFSI lies near the HOMO partition boundary, and LiDFOB displays the largest functional-group dependence of LUMO hosting, spanning solvent-hosted fractions from 97.9\% for CN to 0.4\% for OMe. Salt identity largely shapes the dominant orbital-localization regime, while functionalization further modulates frontier-orbital hosting within mixed regimes. In parallel, total \Li{}--donor bond order responds mainly to salt identity and first-shell topology, whereas the narrower \Li{}-centered ELF deformation reports a distinct reorganization of local electron localization.

Together, these results suggest a hierarchical view: functional groups tune acceptor levels, local fields, and donor character; salt anions influence frontier-state localization in the first shell; and coordination topology modulates aggregate \Li{}--environment contact. These findings add electronic-structure guidance for rational electrolyte design and supply a component for future closed-loop discovery workflows.

\section{Methods}

\subsection{Structure generation and relaxation}

Structures for both libraries are prepared by \emolconf{} and relaxed before electronic-structure evaluation: the DFT-labeled molecular subset with GFN2-xTB\cite{bannwarth_gfn2xtb_2019}, and the \Li{} first-shell structures with the Universal Models for Atoms (UMA) potential\cite{wood_universal_2026} (Supplementary Information, Section~S1). The same scaffold--functional-group combination or the same \Li{} first-shell composition can generate multiple initial structures, to cover different local orientations and metastable coordination arrangements. After relaxation, structures of the same chemical composition are ranked by relative energy, and the relaxed structures are passed to \emoles{} density-matrix inference and \emolpipe{} electronic-structure parsing. The in-house DFT reference labels used for training and benchmarking were generated using Gaussian~16\cite{Gaussian16} at the B3LYP/def2SVP level with GD3 dispersion and a PCM solvent model using the stored dielectric constant; Section~S1 gives the remaining settings.

\subsection{Molecular functionalization}

Each functional group is stored as a three-dimensional template whose terminal marker atom defines its bonding vector, and attachment aligns this vector antiparallel to the parent C--H axis at a covalent-radius-derived bond length, leaving the parent geometry itself untouched. A torsional scan about the nascent bond then selects a sterically clash-free rotamer, and valence checks with layer-wise International Chemical Identifier (InChI) deduplication keep the products unique and chemically valid through exact-depth enumeration of up to four functionalization steps per scaffold.

\subsection{\Li{} first-shell assembly}

For \Li{} first-shell cluster construction, \emolconf{} reads solvents, lithium-salt anions, and local compositions from the Agent-Translated Design Space produced by \emolagent{} and generates initial local structures following the assembly sequence in Figure~\ref{fig:emolconf}. Lewis-basic site mapping first identifies coordination ends that may face \Li{}, including carbonyl, ether, and sulfonyl oxygen, nitrile nitrogen, and fluorine. Initial coordination directions are then placed quasi-uniformly around \Li{} using a Fibonacci sphere, and shell distances are adjusted according to fragment volume to reduce initial overlap when several fragments enter the first shell. Fragment orientations are further refined by constrained rotational sampling so that donor ends maintain reasonable \Li{} contact directions. For sterically congested combinations, adaptive shell expansion increases the shell radius and repeats geometry checks until a relaxable local cluster candidate is obtained. The dielectric constant of each assembled cluster is assigned by similarity matching to a curated 72-solvent library with composition weighting, or fixed to a user-specified value (Supplementary Information, Section~S1.2). Data generation, filtering, and quantum-chemical settings for both libraries are detailed in Section~S1.

\subsection{Density-matrix model}

For density-matrix prediction, \emoles{} is built on the deep-learning tight-binding (DeePTB) framework\cite{Gu2024DeePTB} and its local equivariant representations of quantum operators\cite{Zhouyin2025LEM}, uses SO(2) tensor-product operations inspired by the Equivariant Spherical Channel Network (eSCN)\cite{Passaro2023SO2}, and introduces an electrostatic-aware message-passing (EAMP) module. EAMP applies keep-in-frame SO(2) tensor-product processing: after edge representations are transformed into a bond-aligned local coordinate frame, they remain in the same local reference frame during subsequent message passing and output preprocessing. Keeping edge features in this frame reduces repeated rotations and preserves edge-level electronic-structure information in high-order orbital channels. Because the density matrix is naturally blocked by atom pairs and orbital pairs, \emoles{} moves high-order equivariant self-interaction from the node level to atom-pair and orbital-pair edges, following the higher-body-order motivation of MACE\cite{Batatia2022MACE}. This edge-wise tensor-product self-mixing module is the EACE block introduced above. Before node aggregation, it represents directional charge rearrangement in local coordination environments such as \Li{}--donor and anion--solvent contacts. To balance expressivity and computational cost, this module uses channel-wise tensor products with a uvu connection format and OpenEquivariance acceleration\cite{Bharadwaj2025OpenEquivariance}. At the output end, the element-conditioned tensor-product layer modulates orbital-channel coupling according to atomic element and bond-type scalars, followed by fully connected tensor-product channel mixing before output. This design accommodates element-dependent differences in local charge distributions across multielement chemical environments.

\subsection{Idempotency projection}

The projection is a constrained postprocessing step for closed-shell systems, and the ``i'' in \emolesi{} denotes idempotent. For the density matrix $D^{\mathrm{pred}}_{\mu\nu}$ predicted by \emoles{} and AO overlap matrix $S_{\mu\nu}$, symmetric orthogonalization gives
\begin{equation}
    \tilde{D}^{\mathrm{pred}}
    =
    S^{1/2} D^{\mathrm{pred}} S^{1/2}.
    \label{eq:orthogonal-dm}
\end{equation}
The density matrix in the orthonormal basis is then diagonalized in the natural-orbital basis:
\begin{equation}
    \tilde{D}^{\mathrm{pred}} U
    =
    U n,
    \label{eq:natural-orbital-diagonalization}
\end{equation}
where $U$ is the natural-orbital coefficient matrix and $n$ contains the predicted natural occupations, ordered from largest to smallest with the columns of $U$ reordered accordingly. For a closed-shell system whose total electron number $N_{\mathrm e}$ is determined by chemical composition, the projected occupations are
\begin{equation}
    n_i^{\mathrm{proj}}
    =
    \begin{cases}
        2, & i \leq N_{\mathrm e}/2, \\
        0, & i > N_{\mathrm e}/2.
    \end{cases}
    \label{eq:idempotent-occupations}
\end{equation}
This gives the projected density matrix in the orthonormal basis,
\begin{equation}
    \tilde{D}^{\mathrm{proj}}
    =
    U n^{\mathrm{proj}} U^\mathrm{T}.
    \label{eq:projected-orthogonal-dm}
\end{equation}
Finally, the matrix is transformed back to the nonorthogonal AO basis:
\begin{equation}
    D^{\mathrm{proj}}
    =
    S^{-1/2} \tilde{D}^{\mathrm{proj}} S^{-1/2}.
    \label{eq:projected-dm}
\end{equation}
The projected density matrix $D^{\mathrm{proj}}$, which defines \emolesi{}, therefore satisfies electron-number conservation and the closed-shell idempotency constraint,
\begin{equation}
    \mathrm{Tr}(D^{\mathrm{proj}} S) = N_{\mathrm e},
    \qquad
    D^{\mathrm{proj}} S D^{\mathrm{proj}} = 2D^{\mathrm{proj}}.
    \label{eq:dm-constraints}
\end{equation}

\subsection{Electronic-structure readouts}

\emolpipe{} is the electronic-structure postprocessing and property-parsing layer of \emolstudio{}. It links relaxed geometries, predicted density matrices, and spatial grids, converting model outputs into quantitative electronic-structure readouts with provenance; its electronic-structure and wavefunction-analysis routines build on PySCF, MOKIT, and Multiwfn\cite{Sun2018PySCF,Sun2020PySCF,MOKIT,Lu2012Multiwfn}. The resulting outputs fall into two categories. Scalar descriptors describe the local chemical environment, including HOMO/LUMO energies, gap, dipole moment, ESP extrema, \Li{}--donor bond order, \Li{}-centered ELF deformation factor, and fragment-resolved orbital weights. Spatial electronic-structure objects include HOMO/LUMO isosurfaces, ESP-mapped surfaces, and localized ELF distributions, which are used to analyze frontier-orbital distribution across coordination fragments and to inspect electronic polarization around \Li{}.

For large molecular and cluster libraries, \emolpipe{} exports files compatible with the Visualization for Electronic and Structural Analysis (VESTA)\cite{Momma2011VESTA} and Visual Molecular Dynamics (VMD)\cite{Humphrey1996VMD} programs and parallelizes descriptor extraction through the Portable Batch System (PBS) or DPDispatcher\cite{DPDispatcher2021}. Each output retains provenance links among the structure, density matrix, wavefunction file, cube file, and parsed descriptor, enabling aggregation by parent scaffold, functional-group type, lithium-salt environment, and first-shell topology.

\subsection{Descriptor definitions}

All readouts are evaluated from the same spin-summed one-electron density matrix $D$ in the AO basis\cite{Shao2023DensityMatrix,Rackers2023ElectronDensities}. With AO overlap matrix $S$ and AO functions $\{\chi_\mu\}$, the electron density is
\begin{equation}
\rho_D(\mathbf r)=\sum_{\mu\nu}D_{\mu\nu}\chi_\mu(\mathbf r)\chi_\nu(\mathbf r),
\label{eq:rho-dm}
\end{equation}
and the predicted electron-number error (PENE) introduced above is
\begin{equation}
\PENE=\left|\mathrm{Tr}(D^{\mathrm{pred}}S)-N_{\mathrm e}\right|,
\label{eq:pene}
\end{equation}
reported from the pre-projection matrix, so that it tracks how closely the raw prediction reproduces the correct electron count.

Dipole moments and electrostatic potentials follow from the same density. For Cartesian component $\alpha$,
\begin{equation}
\mu_\alpha(D)=\sum_A Z_A R_{A,\alpha}
-\sum_{\mu\nu}D_{\mu\nu}
\langle\chi_\mu|r_\alpha|\chi_\nu\rangle,
\label{eq:dipole-dm}
\end{equation}
and the electrostatic potential (ESP) at position $\mathbf r$ is
\begin{equation}
V_D(\mathbf r)=
\sum_A\frac{Z_A}{|\mathbf r-\mathbf R_A|}
-\int\frac{\rho_D(\mathbf r')}{|\mathbf r-\mathbf r'|}\,\mathrm{d}\mathbf r'.
\label{eq:esp-dm}
\end{equation}
ESP$_{\max}$ and ESP$_{\min}$ are the extrema of this potential on the sampled molecular surface, describing electron-poor and electron-rich regions and complementing frontier-orbital energies\cite{Peljo2018HomoLumo,Wu2023ESP}.

Frontier-orbital energies are reconstructed by solving the generalized orbital problem from the matrix-conditioned electronic state,
\begin{equation}
F[D]C=SC\varepsilon,
\label{eq:frontier-generalized}
\end{equation}
where $F[D]$ is the Fock matrix constructed from $D$ under the specified electronic-structure protocol, and $C$ and $\varepsilon$ are the orbital coefficients and energies. The highest occupied and lowest unoccupied levels define HOMO, LUMO, and the gap. HOMO and LUMO are interpreted throughout as local frontier-orbital descriptors of the matrix-conditioned electronic state; their relation to observed oxidative or reductive stability depends additionally on solvation, ion pairing, interface potential, and the accessible reaction pathway. Frontier-orbital coefficient similarity is the AO-overlap-weighted coefficient overlap between predicted and reference orbitals,
\begin{equation}
s_k=100\,\left|{\mathbf c}_{k}^{\mathrm{pred}\,T}S{\mathbf c}_{k}^{\mathrm{ref}}\right|.
\label{eq:orbital-similarity}
\end{equation}

Mayer bond order is computed from the same AO density--overlap product. For atoms $A$ and $B$,
\begin{equation}
\mathrm{BO}^{\mathrm{Mayer}}(A,B)=
\sum_{\mu\in A}\sum_{\nu\in B}
(DS)_{\mu\nu}(DS)_{\nu\mu}.
\label{eq:mayer-bo}
\end{equation}
For each \Li{}-centered structure, donor candidates are O, N, F, and S atoms in the first coordination shell, with distance thresholds $\tau_A=2.8$~\AA{} for O, N, and F and $\tau_A=3.2$~\AA{} for S. The threshold selects chemically plausible donor contacts and the Mayer term supplies the \Li{}--donor electronic-contact magnitude. The total \Li{}--donor Mayer bond order is
\begin{equation}
\BOtot
=\sum_{A\in\{\mathrm{O,N,F,S}\}}
\mathrm{BO}^{\mathrm{Mayer}}(\mathrm{Li},A)\,
\mathbf{1}\!\left[r_{\mathrm{Li}A}\leq\tau_A\right],
\label{eq:bo-total}
\end{equation}
and the donor-normalized average is
\begin{equation}
\BOavg=\frac{\BOtot}
{\sum_{A\in\{\mathrm{O,N,F,S}\}}\mathbf{1}\!\left[r_{\mathrm{Li}A}\leq\tau_A\right]}.
\label{eq:bo-avg}
\end{equation}
The benchmark table and the molecular functionalization responses use $\BOavg$, which compares per-donor contact across structures with different donor counts, whereas the cluster motif and salt statistics use $\BOtot$ to describe cumulative first-shell donor strength. This separation keeps model-error comparisons donor-normalized while retaining the shell-level coordination descriptor used in the cluster analysis\cite{Yao2025LithiumBond,Holoubek2022Desolvation}.

Fragment-resolved frontier-orbital weights are Mulliken-like atom-population sums grouped by structure metadata. For orbital $k$ and fragment role $g$,
\begin{equation}
w_{g,k}=100\,\sum_{A\in g}\sum_{\mu\in A}
c_{\mu k}(S\mathbf c_k)_\mu ,
\label{eq:fragment-weight}
\end{equation}
with atom-resolved weights summed into solvent, anion, cation, and unknown roles. The primary solvent--anion coordinate is
\begin{equation}
f_{\mathrm{s|s+a}}=100\,
\frac{w_{\mathrm{candidate\ solvent}}+w_{\mathrm{background\ solvent}}}
{w_{\mathrm{anion}}+w_{\mathrm{candidate\ solvent}}+w_{\mathrm{background\ solvent}}},
\label{eq:host-balance}
\end{equation}
so that 0\% and 100\% denote anion- and solvent-dominated localization conditional on the solvent-plus-anion contribution. A structure is counted as solvent-hosted when $f_{\mathrm{s|s+a}}>50\%$ and anion-hosted otherwise; every solvent-hosted percentage reported in the main text and in the Supplementary Information is the fraction of structures satisfying this condition, not an average of $f_{\mathrm{s|s+a}}$. Cation weight remains available in each record and is summarized separately in the Supplementary Information.

The \Li{}-centered electron localization function (ELF) deformation factor $\phi$ is a surface-based descriptor. The density-matrix-conditioned electronic state is converted to a wavefunction-analysis representation, Multiwfn samples the ELF field $\eta_D(\mathbf r)$ on a common grid, and \emolpipe{} extracts the \Li{}-centered point set
\begin{equation}
\mathcal P_{\mathrm{Li}}=
\left\{\mathbf r_i:\ |\eta_D(\mathbf r_i)-\eta_0|\leq\delta_\eta,\
|\mathbf r_i-\mathbf R_{\mathrm{Li}}|\leq R_{\mathrm{Li}}\right\},
\label{eq:li-elf-pointset}
\end{equation}
which is reduced to the dimensionless deformation factor $\phi$. All structures use the same ELF isovalue band, \Li{}-centered spatial window, and numerical surface construction, so that $\phi$ serves as a comparative localization descriptor.

Finally, every molecular functionalization response reported in this work is a matched difference against the parent scaffold. For any descriptor $X$,
\begin{equation}
\Delta X=X_{\mathrm{functionalized}}-X_{\mathrm{parent}},
\label{eq:delta}
\end{equation}
evaluated with HOMO, LUMO, and gap reconstructed from the closed-shell-projected density matrix, and with ESP extrema, dipole moment, $\BOavg$, and $\phi$ taken from the corresponding descriptor library and parent reference.

\subsection{Design-space translation and interactive analysis}

\emolagent{} is the interaction and physicochemical interpretation layer of \emolstudio{}. It translates a researcher's natural-language question into a structured Agent-Translated Design Space by identifying candidate solvents, salt environments, target first-shell topologies, and requested electronic-structure readouts such as HOMO/LUMO, ESP, \Li{}--donor bond order, orbital localization, and \Li{}-centered ELF deformation. The salt environment, solvent composition, and coordination motif listed at the top of Figure~\ref{fig:emolconf} are one such specification. These structured definitions are dispatched to the integrated \emolconf{}--\emoles{}--\emolpipe{} workflow. During execution, \emolagent{} organizes automated molecular modeling, calls \emolesi{} for density-matrix prediction and idempotency projection, and triggers electronic-structure postprocessing. The resulting physicochemical data are integrated into interactive views containing topology renderings, ESP surfaces, and energy-level comparisons, allowing researchers to evaluate key descriptors and adjust the design space. \emolagent{} also integrates a retrieval-augmented knowledge module linked to curated electrolyte-chemistry and computational-methodology resources, which supports system-specific physical interpretation, parameter provenance, and follow-up analysis suggestions.

\subsection{Benchmarking}

\emoles{} and \emolesi{} are compared with LEM, SLEM, and QHNet\cite{Zhouyin2025LEM,Yu2023QHNet} on held-out sets of 500 molecules and 600 first-shell clusters using density-matrix and electronic-descriptor errors (Table~\ref{tab:emoles-benchmark}). The QH9 test split\cite{Yu2023QH9} provides a complementary benchmark of 9,335 organic molecular structures. Test-set construction is described in Supplementary Information Section~S1; architecture ablations, the QH9 benchmark and the relation between PENE and descriptor accuracy are presented in Section~S2.

\subsection{Statistics and reproducibility}

Functionalization responses are reported as changes relative to the corresponding parent scaffold under a common generation, relaxation, and analysis protocol, so that each reported value is a matched within-scaffold difference rather than a comparison across unrelated molecules. The single-site analysis covers 19,457 matched structures and the multisite analysis 161,081 matched structures across one to four functionalized sites; the first-shell analysis covers 22,500 clusters, 5,625 per salt and 7,500 per coordination motif. Distributions are summarized by medians and quantile ranges rather than by means, except in Figure~\ref{fig:cluster}a, where the per-structure hosting distributions are additionally marked with their means. Structure counts entering each panel, together with the complete summary statistics, are tabulated in Sections~S4 and S5 of the Supplementary Information.

\section*{Data availability}

All datasets generated and analysed in this study will be made publicly available upon publication of the manuscript. These include the functionalized-molecule library of 163,655 structures, the library of 22,500 explicit \Li{} first-shell clusters, the DFT reference labels described in Methods, and the descriptor tables underlying Figures~\ref{fig:smallmol} and~\ref{fig:cluster} and Supplementary Tables~S5--S10.

\section*{Code availability}

The \emolstudio{} source code is provided as Supplementary Software~1. The archive includes \emolconf{}, the \emoles{} inference interface, \emolpipe{}, example checkpoints and structures, and scripts for inference, postprocessing and table reproduction. Installation instructions and a detailed reproducibility guide accompany the repository.

\section*{Acknowledgements}

We thank Prof. Xiang Chen of the Department of Chemical Engineering, Tsinghua University, for helpful discussions on electrolyte chemistry and for comments on the manuscript. The authors acknowledge the High Performance Computing facilities at the National University of Singapore (NUS HPC), under Project CFP04-CF-019, for computational resources and access to licensed software.

\section*{Funding}

This work was supported by the Singapore Ministry of Education Tier 2 grant (No. MOE-T2EP50123-0012) to L.S.

\section*{Author contributions}

M.L. implemented the \emolstudio{} framework, performed the calculations and analysis, and wrote the original draft. H.Y. developed the \emolagent{} module. L.S. supervised the project and acquired funding. All authors reviewed and approved the final version of the manuscript.

\section*{Competing interests}

The authors declare no competing interests.

\endgroup
\clearpage
\newgeometry{left=2cm,right=2cm,top=2cm,bottom=2cm}
\setstretch{1}
\fontsize{10}{12}\selectfont
\let\addcontentsline\savedaddcontentsline
\setcounter{section}{0}
\setcounter{subsection}{0}
\setcounter{figure}{0}
\setcounter{table}{0}
\setcounter{equation}{0}
\setcounter{page}{1}
\renewcommand{\thesection}{S\arabic{section}}
\renewcommand{\thefigure}{S\arabic{figure}}
\renewcommand{\thetable}{S\arabic{table}}
\renewcommand{\theequation}{S\arabic{equation}}
\renewcommand{\thepage}{S\arabic{page}}
\renewcommand{\theHsection}{si.\arabic{section}}
\renewcommand{\theHequation}{si.\arabic{equation}}
\renewcommand{\theHfigure}{si.\arabic{figure}}
\renewcommand{\theHtable}{si.\arabic{table}}
\setcounter{secnumdepth}{3}
\setcounter{tocdepth}{2}
\renewcommand{\contentsname}{Table of Contents}

\captionsetup{font=small,labelfont=bf,labelsep=period,hypcap=false}
\setlength{\parskip}{0.18em}
\setlength{\tabcolsep}{4.5pt}
\renewcommand{\arraystretch}{1.12}
\pdfbookmark[0]{Supplementary Information}{si-part}
\begingroup

\begin{center}
{\Large\bfseries Supplementary Information for\par}
\vspace{0.5em}
{\large\bfseries A Density-Matrix Framework for Electronic-Structure Analysis of Electrolytes for Lithium Batteries\par}
\vspace{1em}

\begin{minipage}{0.97\textwidth}
\centering
\normalsize
Mingkang Liu\textsuperscript{1},
Huize Yu\textsuperscript{1}
and Lei Shen\textsuperscript{1,2,3,*}\\[0.7em]
\footnotesize\itshape
\textsuperscript{1}Department of Mechanical Engineering, National University of Singapore, 9 Engineering Drive 1, Singapore 117575, Singapore\\
\textsuperscript{2}National University of Singapore (Chongqing) Research Institute, Chongqing 401120, China\\
\textsuperscript{3}National University of Singapore Guangzhou Research Translation and Innovation Institute, Guangzhou, Guangdong 510700, China\\[0.55em]
\upshape
\textsuperscript{*}Corresponding author. E-mail: shenlei@nus.edu.sg (L.S.)
\end{minipage}%
\end{center}
\vspace{-0.6em}

\begin{center}
\footnotesize Pages S1--\pageref{LastPage}\quad
Figures S1--S7\quad Tables S1--S10
\end{center}
\vspace{0.4em}

\tableofcontents
\newpage

\section{Dataset Construction and Computational Details}
\label{si-sec:data-scope}

\emolconf{} prepares the two structure libraries analyzed in the main text: a parent-matched functionalized-molecule library and an explicit \Li{} first-shell cluster library. For each, a DFT-labeled subset provides reference data for model training and evaluation, and a larger descriptor library (163,655 molecular products and 22,500 four-salt clusters) supports the statistical analyses. Table~\ref{si-tab:scope} summarizes the analysis sets and the sample counts on which each main-text statistic is evaluated, and Table~\ref{si-tab:emolconf-lineage} lists the source sets and their roles in this work.

\begin{table}[htbp]
  \centering
  \small
  \caption{Analysis sets, sample counts, and reporting conventions. ``Paired-valid'' means that a functionalized structure and its corresponding parent both contain the entries needed to evaluate the parent-referenced response.}
  \label{si-tab:scope}
  \begin{tabularx}{\textwidth}{>{\bfseries}p{0.16\textwidth}p{0.27\textwidth}p{0.25\textwidth}X}
    \toprule
    Analysis set & Source or construction scope & Reported sample counts & Reporting convention \\
    \midrule
    Molecular functionalization & 1,399 parents; 163,655 functionalized molecules; eight functional-group classes; one to four functionalized sites & 161,081 paired-valid structures; 19,457 single-site structures & Parent-referenced gas-phase convention; projected closed-shell HOMO/LUMO/gap; density-matrix electron-count consistency retained as a descriptor \\
    Explicit first shell & 22,500 relaxed \Li{}-centered clusters; four salts; three fixed motifs & 22,500 structures; 5,625 per salt; 7,500 per motif; 1,875 per salt--motif cell & Relaxed cluster library with projected closed-shell readout \\
    Model reference tests & In-house molecule and cluster test sets drawn from the DFT-labeled reference data & 500 molecules and 600 clusters per model & Matched structures and common descriptor definitions \\
    QH9 comparison & Original QH9 test split; density matrices and downstream descriptor labels recomputed under the common protocol & 9,335 structures & Raw/projected comparison on matched structures \\
    \bottomrule
  \end{tabularx}
\end{table}

Each structure in these libraries is stored as a structured record so that every statistic can be traced back to its origin. For a molecule, the record links the parent identifier, functionalization site, functional-group class, functionalization count, charge and spin state, geometry identifier, matrix state, and descriptor status. For a cluster, it links the salt, candidate solvent, optional background solvent, coordination motif, fragment-role map, charge and spin state, source-structure identifier, pre-projection PENE (Methods), projection status, and parsed descriptors. Because composition and fragment roles are stored explicitly, orbital contributions can be aggregated using consistent atom-to-fragment assignments.

\begin{table}[htbp]
  \centering
  \small
  \caption{\emolconf{} source sets and their roles. The functionalized-molecule construction and the 7,200-cluster reference set define structures and DFT labels; the molecular and four-salt cluster statistics use the larger analysis-scale descriptor libraries.}
  \label{si-tab:emolconf-lineage}
  \begin{tabularx}{\textwidth}{>{\bfseries}p{0.20\textwidth}p{0.28\textwidth}p{0.24\textwidth}X}
    \toprule
    Source set & Object scope & Count & Use in this work \\
    \midrule
    Literature CHO parents & Gao et al. electrolyte solvent parents\cite{SI-Gao2023JACS}; C/O frameworks with heavy-atom count no larger than 9 & 1,399 parents and 1,399 matched \Li{}-coordinated parents & Parent scaffold source for matched molecular response \\
    Functionalized DFT pairs & Eight functional-group fragments attached to editable sites; neutral and \Li{}-coordinated structures follow the same functionalization sequence & 4,070 candidates; 4,045 xTB-optimized structures per charge state; 4,020 Gaussian-paired neutral/\Li{} products & Molecular pair set for functionalization and \Li{} coordination \\
    Cluster DFT reference set & 72 solvents, 10 anions, and 10 \Li{} first-shell coordination modes spanning solvent-separated ion-pair (SSIP), contact ion-pair (CIP), and aggregate (AGG) environments & 7,200 DFT-labeled entries; train/valid/test = 6,000/600/600 & Cluster label set for model evaluation \\
    Four-salt analysis clusters & Functionalized solvent library combined with LiPF$_6$, LiFSI, LiDFOB, and LiTDI under M$_2$X, MBX, and MX$_2$ motifs & 22,500 relaxed structures & Salt-resolved library for cluster chemistry statistics \\
    \bottomrule
  \end{tabularx}
\end{table}

\subsection{Functionalized-Molecule Library}
\label{si-sec:molecule-construction}

The molecular lineage starts from 1,399 CHO electrolyte solvent parents reported by Gao et al.\cite{SI-Gao2023JACS}. That parent library was generated from formaldehyde and dimethyl ether fragments by graph-based enumeration, limited to C/O frameworks with no more than 9 heavy atoms, and filtered to remove alcohol and acid structures with labile hydrogens. These neutral parents are paired with corresponding \Li{}-coordinated parent structures so that intrinsic molecular descriptors and \Li{}-coordination descriptors carry a common scaffold identifier.

\emolconf{} performs functionalization as a constrained three-dimensional structure edit (Figure~\ref{si-fig:emolconf-molecule-samples}). The procedure builds a covalent-radius connectivity graph, identifies editable C--H sites, treats O--H sites separately when present, and attaches one of eight chemically defined fragments--methyl (Me), cyano (CN), trifluoromethyl (CF$_3$), fluoro (F), methoxy (OMe), methoxycarbonyl (CO$_2$Me), methylsulfonyl (SO$_2$Me), and fluorosulfonyl (SO$_2$F). Each fragment carries an attachment marker at the connection site. During insertion the marker is removed, the fragment is rotated onto the parent heavy-atom--H direction, the initial bond length is set from covalent radii, and torsions around the new bond are scanned to avoid short nonbonded contacts. RDKit bond perception, SMILES/InChI identity checks, single-component filtering, small-ring exclusion, abnormal-bond filtering, and layer-wise InChI deduplication keep the products inside the target chemical space.

\begin{figure}[htbp]
  \centering
  \includegraphics[width=\linewidth]{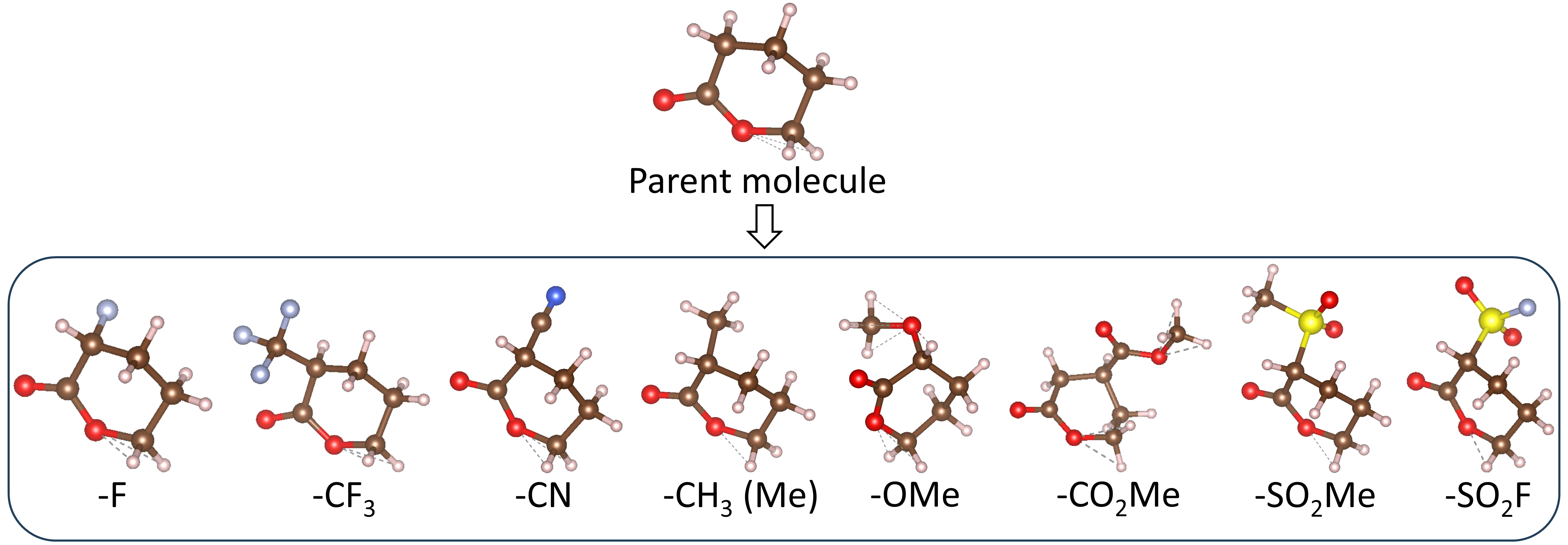}
  \caption{\textbf{Representative \emolconf{} site-level functional-group editing.} Functional-group fragments are rigidly attached to an editable site of the parent scaffold while preserving the parent conformation, producing the corresponding functionalized structures used to calculate parent-referenced molecular responses. Atom colors in this and all subsequent atomistic renderings: Li light green, C brown, O red, H white, N blue, S yellow, F pale blue, P violet, B pink, Cl green.}
  \label{si-fig:emolconf-molecule-samples}
\end{figure}

Neutral and \Li{}-coordinated products follow the same functionalization sequence: parent scaffold, edited-site sequence, number of functionalized sites, and functional-group sequence are replayed in the \Li{}-coordinated structure. This DFT-labeled subset comprises 4,070 neutral candidates and 4,070 matched \Li{}-coordinated candidates, consisting of 2,697 single-group-route and 1,373 mixed-route products. Counts by number of functionalized sites were $n=1$: 671, $n=2$: 2,040, $n=3$: 673, and $n=4$: 686. After xTB geometry optimization\cite{SI-bannwarth_gfn2xtb_2019}, 4,045 neutral and 4,045 \Li{}-coordinated structures remained; the Gaussian-paired intersection contained 4,020 neutral/\Li{} pairs covering 1,373 parent molecules.
The in-house molecule test set used for the model benchmarks comprises 500 of these structures (Table~\ref{si-tab:scope}), held out from the data used to fit the models. The molecular statistics in the main text and in Section~\ref{si-sec:molecules} summarize the full descriptor library: 163,655 generated products, 161,081 paired-valid parent-referenced responses, and 19,457 single-site responses.

\subsection{Explicit \Li{} First-Shell Cluster Libraries}
\label{si-sec:cluster-construction}

The DFT-labeled cluster reference set spans 72 electrolyte solvents and 10 lithium-salt anions: BF$_4^-$, FSI$^-$, PF$_6^-$, TDI$^-$, TFSI$^-$, ClO$_4^-$, NO$_3^-$, DFOB$^-$, BOB$^-$, and PO$_2$F$_2^-$ (Figure~\ref{si-fig:emolconf-cluster-samples}). The solvent entries carry dielectric constants collected from publicly available sources, which are passed to the Gaussian PCM setting. At inference time, the dielectric constant assigned to a cluster is derived from this library: each solvent molecule in the cluster is matched by molecular similarity to its closest library entry, and the dielectric constants of the matched entries are combined as a composition-weighted average over the solvent molecules in the cluster. An interface is also provided to fix the dielectric constant to a user-specified value in place of this similarity-based assignment. All cluster downstream-task calculations reported in this work take the dielectric-dependent PCM environment into account. For every solvent--anion combination, 10 \Li{} first-shell coordination modes are enumerated: solvent-separated ion-pair (SSIP) structures with no anion in the first shell and total coordination number 1--4, contact ion-pair (CIP) structures with one anion and total coordination number 2--4, and aggregate (AGG) structures with two anions and total coordination number 3--4 or three anions and total coordination number 4. The 7,200 DFT-labeled entries were organized on a unified 72-solvent $\times$ 10-salt $\times$ 10-coordination grid. Because SSIP clusters do not contain a salt anion, the salt-index dimension was retained for compatibility with this unified indexing scheme. For these SSIP entries, the repeated salt indices correspond to independent initial geometries generated by the stochastic \emolconf{} cluster-assembly procedure rather than to distinct chemical compositions. These geometries sample microscopic structural perturbations under the same composition. After removing this SSIP indexing redundancy, the 7,200 labeled entries correspond to 4,608 unique structures, with the SSIP subset reduced from 2,880 indexed entries to 288 unique structures, alongside 2,160 CIP and 2,160 AGG clusters.

\begin{figure}[htbp]
  \centering
  \includegraphics[width=0.92\linewidth]{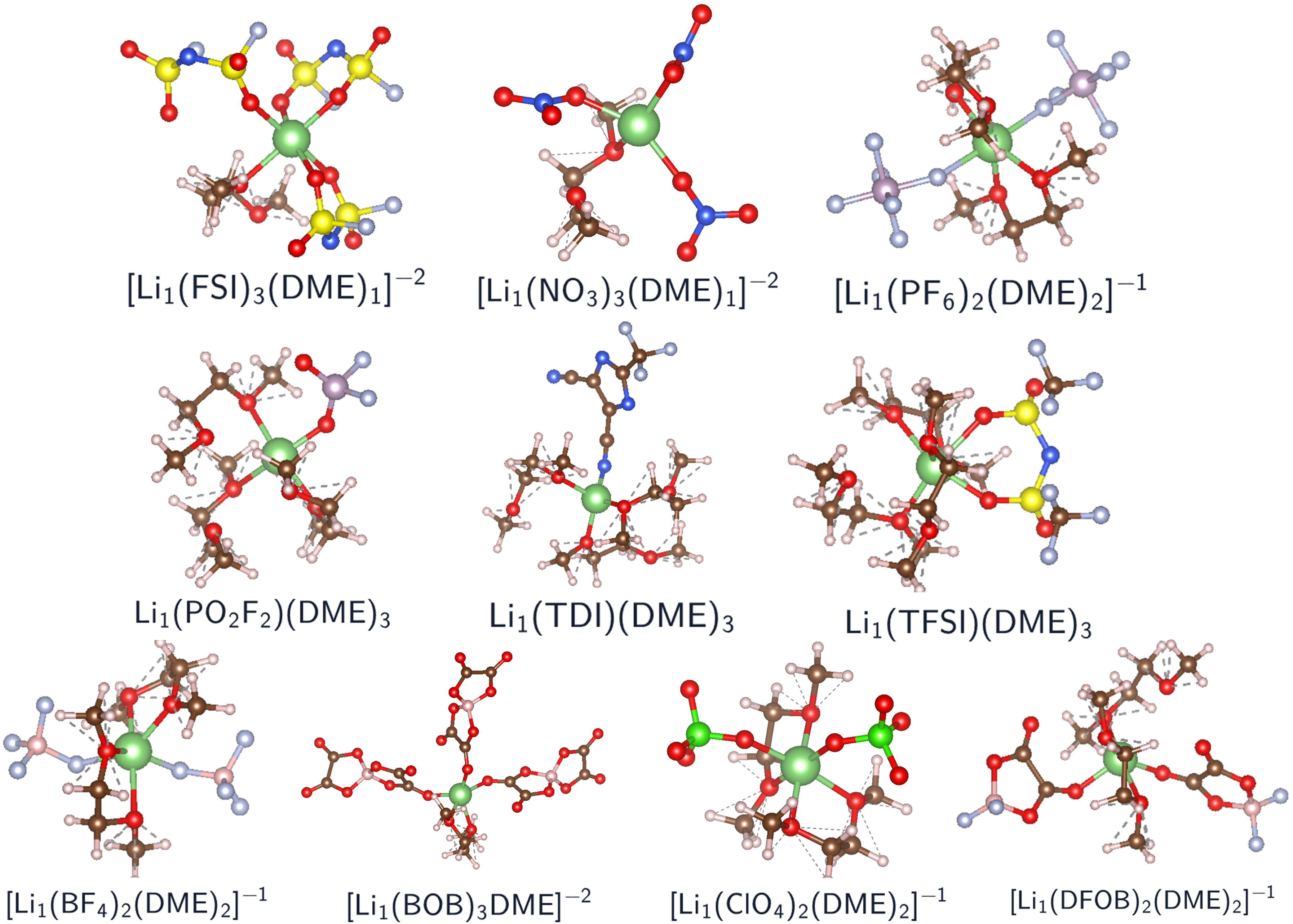}
  \caption{\textbf{Assembled \Li{}-centered first-shell clusters across the 10-anion reference set.} 1,2-Dimethoxyethane (DME) is the common solvent, with one cluster shown for each of the 10 lithium-salt anions in the DFT-labeled reference set; panel labels give the cluster composition and net charge.}
  \label{si-fig:emolconf-cluster-samples}
\end{figure}

All 7,200 reference structures are relaxed with the Universal Models for Atoms (UMA) potential and then evaluated with Gaussian~16 at the B3LYP/def2SVP level with GD3 dispersion, a generic PCM solvent model using the stored dielectric constant, and full population/NBO output. The Gaussian outputs are converted with \texttt{formchk}; Hamiltonian, overlap, density matrix, highest occupied molecular orbital (HOMO), lowest unoccupied molecular orbital (LUMO), and gap labels are parsed from the resulting electronic-structure outputs. The label representation uses a common def2SVP AO convention for Li/C/N/S/O/F/H/B/Cl/P so that Hamiltonian and density-matrix blocks have a consistent ordering. The resulting label set contains all 7,200 entries and is split into non-overlapping train/valid/test partitions of 6,000, 600, and 600 structures.

The four-salt analysis library assembles local solvation environments by combining \Li{}, one salt anion, candidate solvent molecules from the functionalized-molecule library, and, where required, a background solvent. The four salts are LiPF$_6$, LiFSI, LiDFOB, and LiTDI. Three fixed first-shell motifs are generated for each salt: solvent-rich M$_2$X with two candidate solvents and one anion, diluent-containing MBX with one candidate solvent, one background solvent, and one anion, and anion-rich MX$_2$ with one candidate solvent and two anions. Initial \Li{}-centered structures are relaxed before electronic-structure readout, and the fragment-role map defines solvent, anion, cation, and background-solvent orbital partitions. The resulting descriptor library contains 22,500 relaxed structures, with 5,625 per salt, 7,500 per motif, and 1,875 per salt--motif cell. The cluster statistics in the main text and in Section~\ref{si-sec:clusters} summarize this 22,500-structure library; descriptor readouts are obtained from the matrix-conditioned electronic state, and PENE is reported separately from the corresponding pre-projection density matrix.

\FloatBarrier
\section{Supplementary Evaluation of \emoles{}}
\label{si-sec:emoles-supplement}

This section collects model-level evidence complementing the benchmark in the main text: an ablation of the \emoles{} architecture components, a benchmark on the QH9 test split, and an analysis of PENE as an inference-time confidence indicator.

\subsection{Contribution of the Architecture Components}
\label{si-sec:ablation}

Table~\ref{si-tab:ablation} isolates the contribution of each \emoles{} architecture component on the cluster test set.

\begin{table}[H]
  \centering
  \footnotesize
  \setlength{\tabcolsep}{3pt}
  \caption{Cluster-set component ablation using the unprojected \emoles{} processing protocol in Table~1 of the main manuscript. Relative changes are referenced to the unprojected \emoles{} baseline. DM-MAE is reported in $10^{-4}$, Hamiltonian MAE in $10^{-6}$ Ha, dipole in Debye, PENE in electrons, and \Li{} ELF deformation factor and \Li{} avg donor BO in $10^{-3}$. \Li{} avg donor BO denotes the average \Li{}--donor electronic-contact Mayer bond-order MAE.}
  \label{si-tab:ablation}
  \begin{tabular}{lrrrrrrrr}
    \toprule
    Variant & \makecell{DM-MAE\\($10^{-4}$)} & \makecell{$\Delta$DM\\(\%)} & \makecell{Ham-MAE\\($10^{-6}$ Ha)} & \makecell{$\Delta$Ham\\(\%)} & \makecell{Dipole\\(D)} & \makecell{PENE\\(e)} & \makecell{Li ELF def.\\($10^{-3}$)} & \makecell{Li avg\\donor BO\\($10^{-3}$)} \\
    \midrule
    \emoles{} baseline & 3.26 & 0.0 & 69.5 & 0.0 & 0.241 & 0.0275 & 0.743 & 0.578 \\
    no bond frame & 3.34 & +2.5 & 73.0 & +5.0 & 0.256 & 0.0294 & 0.853 & 0.599 \\
    no edge self-mix & 3.64 & +11.7 & 78.6 & +13.1 & 0.272 & 0.0306 & 0.779 & 0.671 \\
    no output tensor product & 4.08 & +25.2 & 94.7 & +36.3 & 0.346 & 0.0378 & 0.894 & 0.879 \\
    no all three & 4.45 & +36.5 & 101.0 & +45.3 & 0.346 & 0.0396 & 0.994 & 0.905 \\
    \bottomrule
  \end{tabular}
\end{table}

The smallest individual penalty comes from removing the bond-aligned frame, which mainly preserves the orientation of AO-pair information along a coordination contact. Removing edge-wise self-mix gives a larger loss, indicating that \Li{}--donor and anion--solvent pair features benefit from local many-body mixing before they enter the final atom-pair blocks. The element-conditioned output layer gives the largest individual contribution, consistent with the multielement nature of salt clusters and the need to distinguish element-pair density-matrix blocks. Each ablation increases the \Li{} ELF deformation-factor and average \Li{}--donor BO errors relative to the full model, although their rankings differ across descriptors. Combined removal gives the largest errors for both local-environment descriptors, supporting the joint use of the three components.

\subsection{Benchmark on the QH9 Test Split}
\label{si-sec:qh9-check}

The QH9 test split comprises 9,335 organic molecular structures, with reference density matrices, orbital energies and descriptors recomputed using PySCF. Table~\ref{si-tab:qh9-downstream} compares five model variants on this common reference set.

\begin{table}[htbp]
  \centering
  \small
  \caption{Downstream descriptor errors on the QH9 test split ($N=9{,}335$). Orbital and ESP mean absolute errors are in eV; frontier-orbital coefficient similarity values are percentages. The suffix -i denotes idempotency projection.}
  \label{si-tab:qh9-downstream}
  \begin{tabular}{lrrrrrrr}
    \toprule
    Model & HOMO & LUMO & Gap & HOMO sim. & LUMO sim. & ESP$_{\max}$ & ESP$_{\min}$ \\
    \midrule
    \emolesI{} & 0.0143 & 0.0057 & 0.0118 & 99.91 & 99.99 & 0.0047 & 0.0068 \\
    \emoles{}  & 0.1686 & 0.1256 & 0.0569 & 99.22 & 99.64 & 0.0205 & 0.0272 \\
    LEM         & 0.2020 & 0.1516 & 0.0697 & 98.87 & 99.60 & 0.0225 & 0.0281 \\
    SLEM        & 0.2188 & 0.1610 & 0.0786 & 98.47 & 99.43 & 0.0273 & 0.0343 \\
    QHNet-i     & 0.0229 & 0.0098 & 0.0191 & 99.81 & 99.96 & 0.0068 & 0.0107 \\
    \bottomrule
  \end{tabular}
\end{table}

Projection mainly improves descriptors that depend on a chemically well-defined occupied manifold. Across the 9,335 QH9 structures, projection of the \emoles{} predictions lowers the density-matrix MAE from $4.94$ to $4.62\times10^{-4}$, restores electron counting from an error of $2.4\times10^{-2}$ to $3\times10^{-14}$ electrons, and reduces the dipole MAE from $0.315$ to $0.038$~D; local frontier-orbital and ESP readouts become substantially tighter on the same structures. On this test split, \emolesI{} gives the lowest error and the highest orbital similarity for every descriptor in Table~\ref{si-tab:qh9-downstream}.

\subsection{Electron-Count Self-Consistency as a Confidence Indicator}
\label{si-sec:pene-confidence}

PENE, defined in Methods, quantifies density-matrix electron-count self-consistency and accompanies downstream property extraction. The cluster test set is plotted by per-structure PENE and absolute descriptor error,
\begin{equation}
e_X=\left|X^{\mathrm{pred}}-X^{\mathrm{ref}}\right|,
\label{si-eq:descriptor-abs-error}
\end{equation}
where $X$ is HOMO, LUMO, ESP$_{\min}$, or ESP$_{\max}$. The plotted range is $0\leq\PENE\leq0.2$ electrons and contains all 600 cluster-test structures.

The colored summaries in Figure~\ref{si-fig:pene-confidence} are PENE-quantile bins. For each descriptor, structures are sorted by PENE and divided into ten equal-count bins using quantile cuts. Within each bin, the plotted $x$ position is the median PENE and the connected point is the median descriptor error. The darker vertical interval spans the 25th--75th percentiles of $e_X$, and the lighter interval spans the 10th--90th percentiles. The dashed line is a least-squares fit to the binned median trend; gray points show item-level scatter.

Two trend metrics are reported in the figure. Spearman is the rank correlation between PENE and descriptor error,
\begin{equation}
\rho_{\mathrm S}=\mathrm{corr}\!\left(\mathrm{rank}(\PENE),\mathrm{rank}(e_X)\right),
\label{si-eq:pene-spearman}
\end{equation}
and measures monotonic ordering. Bin $R^2$ is computed from the binned medians,
\begin{equation}
R^2_{\mathrm{bin}}=
1-\frac{\sum_j(\tilde e_{X,j}-\hat e_{X,j})^2}
{\sum_j(\tilde e_{X,j}-\bar{\tilde e}_X)^2},
\label{si-eq:pene-bin-r2}
\end{equation}
where $\tilde e_{X,j}$ is the median error in PENE bin $j$ and $\hat e_{X,j}$ is the fitted dashed-line value. Bin $R^2$ describes the smoothness of the median error trend after PENE binning, while the gray points retain the item-level dispersion.

Figure~\ref{si-fig:pene-confidence} shows the same qualitative behavior for all four descriptors: the binned median error is smallest and nearly flat over the low-PENE bins, where most of the test set resides, and rises monotonically as PENE grows. The ordering is strongest for the frontier-orbital energies (Spearman rank correlations of 0.53 and 0.49 for HOMO and LUMO, with binned-trend $R^2$ of 0.98 and 0.90) and somewhat softer for the ESP extrema (0.39--0.40, with $R^2$ of 0.92--0.94), consistent with orbital energies being the readouts most sensitive to an inconsistent electron count, whereas the surface-sampled ESP extrema integrate over the density and respond more gradually. The quantile intervals widen in the same direction, associating larger PENE with both a larger typical error and a broader error spread. Because PENE is computed from the predicted density matrix alone, it provides a label-free, inference-time confidence indicator; structures with large values can be selected for additional checking or reference calculation.

\begin{figure}[H]
  \centering
  \includegraphics[width=\textwidth]{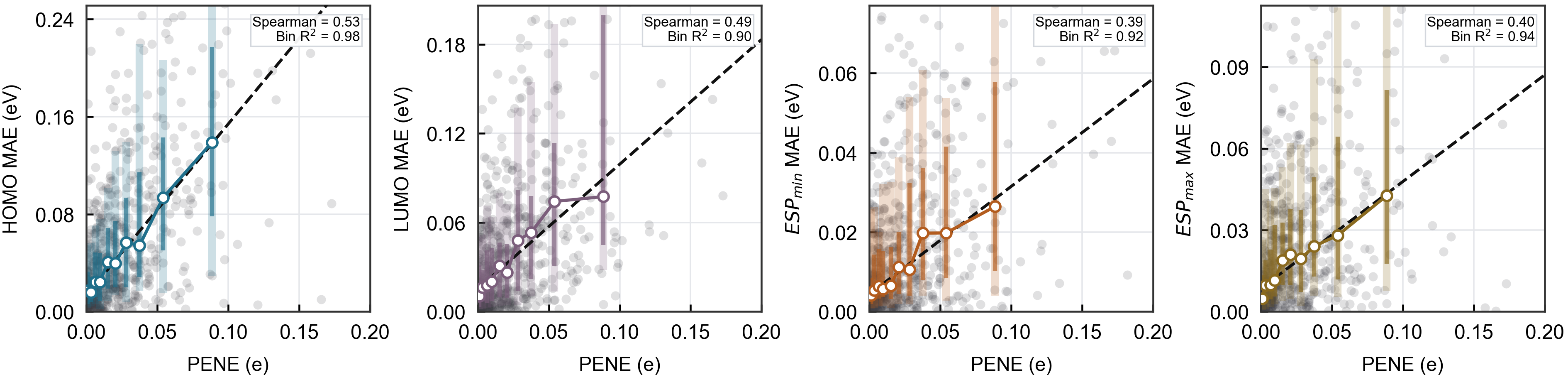}
  \caption{\textbf{PENE-binned cluster descriptor-error trends.} Absolute HOMO, LUMO, ESP$_{\min}$, and ESP$_{\max}$ errors on the 600-structure cluster test set. Gray points are individual structures; colored points and intervals summarize PENE-quantile bins, and panel insets give the Spearman rank correlation and the binned-trend $R^2$.}
  \label{si-fig:pene-confidence}
\end{figure}

\FloatBarrier
\section{\Li{} Ligand Exchange along a Two-Stage Coordination Path}
\label{si-sec:li-two-stage-binding}

To follow a chemically resolved coordination event, we examine \Li{} ligand exchange in which a donor oxygen of an entering bis(fluorosulfonyl)imide (FSI$^-$) anion joins the first coordination shell of a DME-solvated \Li{} while a DME donor oxygen departs. Separating donor approach from ligand release, a two-stage exchange coordinate is defined over 1000 frames (Figure~\ref{si-fig:li-two-stage}a). Frames 0--500 describe FSI$^-$ approach: the entering Li--O(FSI$^-$) contact contracts from 4.20 to 2.00~\AA{}, while the leaving Li--O(DME) contact remains close to 2.07~\AA{}. Frames 500--1000 describe DME release: the newly formed Li--O(FSI$^-$) contact is retained in the first coordination shell, while the Li--O(DME) contact is elongated to 4.20~\AA{}. This construction keeps the two geometric operations explicit while providing a single coordinate for energetic and electronic-structure readouts.

\begin{figure}[htbp]
  \centering
  \includegraphics[width=\linewidth]{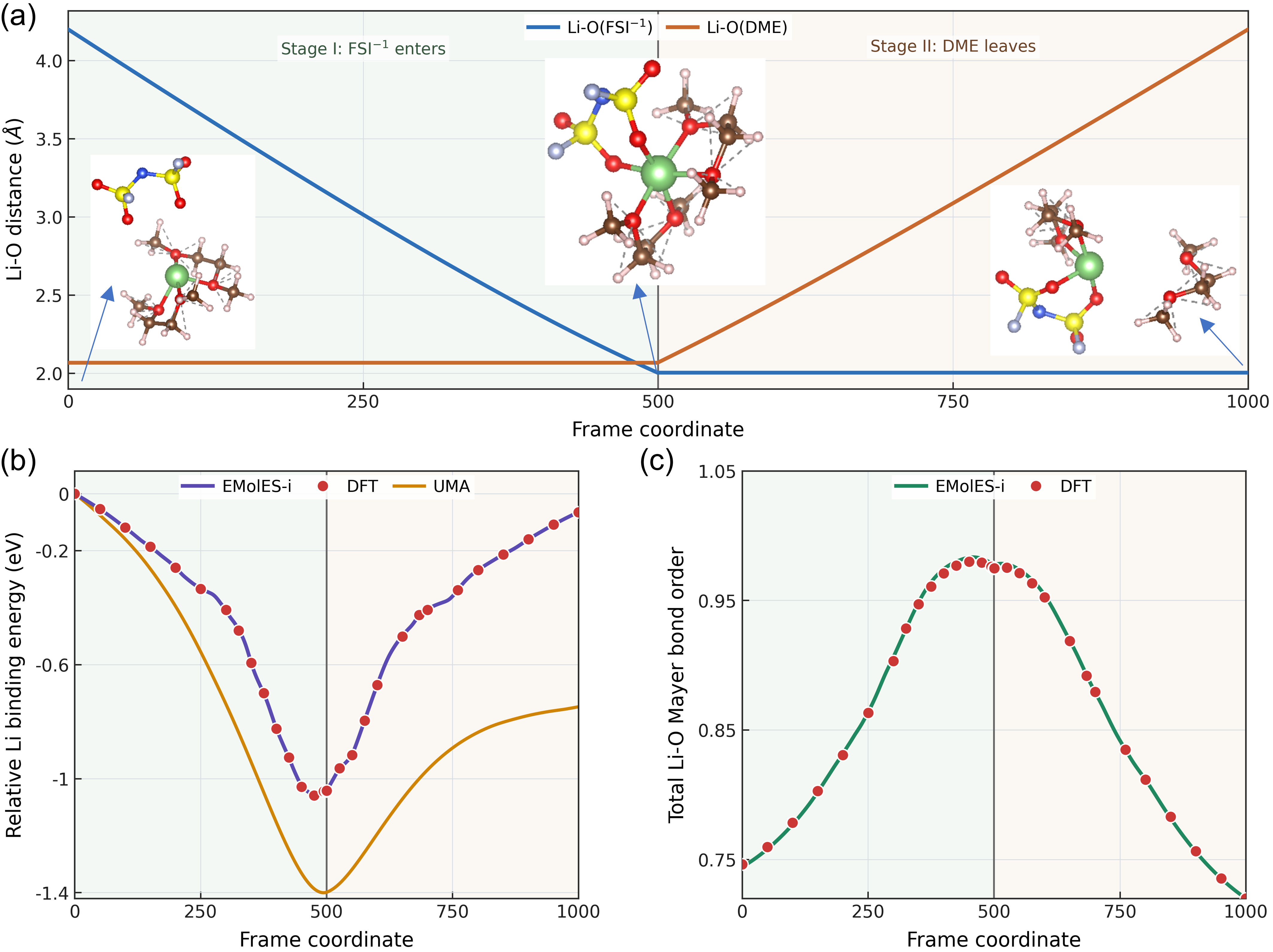}
  \caption{\textbf{\Li{} ligand exchange along the two-stage coordination path.} (a) Two-stage exchange coordinate with representative structures: the entering Li--O(FSI$^-$) contact (blue) closes from 4.20 to 2.00~\AA{} over frames 0--500 (Stage I), and the leaving Li--O(DME) contact (orange) is released to 4.20~\AA{} over frames 500--1000 (Stage II); insets show snapshots at the start of the path, the frame-500 crossover, and the end of the path. (b) Relative \Li{} binding energy computed with the Li-removed ligand framework as the common reference; \emolesI{} denotes the continuous 1000-frame density-matrix profile, red markers are the 30 DFT validation frames, and UMA is shown under the same reference convention. (c) Total Mayer bond order over the six dominant Li--O donor contacts, evaluated from the density matrix and benchmarked against the same DFT frames.}
  \label{si-fig:li-two-stage}
\end{figure}

The relative \Li{} binding energy is evaluated with the Li-removed ligand framework as the common reference at each fixed geometry, so that \emolesI{}, UMA, and DFT are compared under the same convention (Figure~\ref{si-fig:li-two-stage}b). \emolesI{} resolves the full 1000-frame path directly from predicted density matrices, and DFT is evaluated at 30 reference frames to validate the profile. At those frames, \emolesI{} reproduces the DFT binding energies with a mean absolute error of 0.084~eV and a Pearson correlation coefficient of 0.995. UMA shows a larger deviation from DFT under the same reference convention, with a mean absolute error of 0.380~eV and $r=0.888$. The \emolesI{} profile becomes progressively more stabilizing during donor approach, reaches its minimum near the frame-500 crossover, and weakens as the leaving ligand is pulled away.

The local electronic response follows the same sequence. The total Mayer bond order over the six dominant Li--O donor contacts, likewise resolved over all 1000 frames, tracks the DFT values closely at the 30 validation frames, with a mean absolute error of 0.00165 and $r=0.99995$ (Figure~\ref{si-fig:li-two-stage}c). The bond order rises as the entering donor joins the first coordination shell, peaks near the strongest binding region, and decreases during ligand release. Together, the energy and bond-order profiles associate transient coordination-shell closure with the strongest \Li{} binding along this path.

\FloatBarrier

\section{Supplementary Statistics for Functionalized Molecules}
\label{si-sec:molecules}

All molecular responses are parent-referenced shifts $\Delta X$ evaluated on corresponding scaffolds (Methods). The single-site set contains 19,457 paired-valid structures across CN, SO$_2$F, SO$_2$Me, CO$_2$Me, OMe, CF$_3$, F, and Me, and the full paired-valid library contains 161,081 parent-referenced responses.

\subsection{Single-Site Medians and Multi-Site Endpoints}
\label{si-sec:molecule-summary-tables}

Tables~\ref{si-tab:mono} and~\ref{si-tab:parent-esp} collect the summary statistics invoked quantitatively in the main text: parent-referenced medians at one and at four functionalized sites, and parent-conditioned $\Delta$ESP$_{\max}$ endpoint medians.

\begin{table}[H]
  \centering
  \footnotesize
  \setlength{\tabcolsep}{3pt}
  \renewcommand{\arraystretch}{1.08}
  \caption{Parent-referenced functionalization medians at one and at four functionalized sites.
  Energy and ESP shifts are in eV, average \Li{}--donor BO ($\BOavg$) is dimensionless, and dipole
  shifts are in Debye. $N$ is the single-site sample size. Dashes mark endpoints not reported here.}
  \label{si-tab:mono}
  \begin{tabular}{@{}lrrrrrrrrrr@{}}
    \toprule
    & & \multicolumn{2}{c}{$\Delta$HOMO} & \multicolumn{2}{c}{$\Delta$LUMO}
    & \multicolumn{2}{c}{$\Delta\BOavg$} & & & \\
    \cmidrule(lr){3-4}\cmidrule(lr){5-6}\cmidrule(lr){7-8}
    Group & $N$ & $n=1$ & $n=4$ & $n=1$ & $n=4$ & $n=1$ & $n=4$
    & \makecell{$\Delta$ESP$_{\max}$\\($n=1$)} & \makecell{$\Delta$ESP$_{\min}$\\($n=1$)} & \makecell{$\Delta\mu$\\($n=1$)} \\
    \midrule
    CN & 2621 & -0.406 & -1.501 & -0.636 & -2.482 & -0.0073 & -0.0686 & +0.302 & +0.064 & +2.81 \\
    SO$_2$F & 2633 & -0.479 & -1.368 & -0.697 & -2.461 & -0.0189 & -0.1390 & +0.808 & +0.180 & +7.96 \\
    SO$_2$Me & 2619 & -0.154 & -- & -0.445 & -1.569 & -0.0332 & -0.1286 & +0.821 & -0.144 & +4.71 \\
    CO$_2$Me & 2574 & -0.063 & -0.075 & -0.983 & -1.537 & -0.0040 & -0.0556 & +0.202 & -0.030 & +1.69 \\
    OMe & 2225 & +0.080 & +0.320 & -0.014 & -0.077 & +0.0011 & -- & +0.040 & -0.010 & +1.22 \\
    CF$_3$ & 2520 & -0.259 & -0.881 & -0.173 & -0.572 & -0.0053 & -- & +0.227 & +0.172 & +1.43 \\
    F & 2613 & -0.190 & -0.757 & -0.100 & -0.359 & -0.0028 & -- & +0.149 & +0.117 & +1.21 \\
    Me & 1652 & +0.031 & -- & +0.010 & -- & +0.0033 & -- & -0.017 & +0.029 & +0.74 \\
    \bottomrule
  \end{tabular}
\end{table}

\begin{table}[htbp]
  \centering
  \small
  \caption{Parent-conditioned $\Delta$ESP$_{\max}$ endpoint medians. ``Low'' and ``high'' refer to the first and last parent-ESP$_{\max}$ bins used in the molecular analysis. Values are in eV.}
  \label{si-tab:parent-esp}
  \begin{tabular}{lrrrrrrrr}
    \toprule
    Parent bin & CN & SO$_2$F & SO$_2$Me & CO$_2$Me & OMe & CF$_3$ & F & Me \\
    \midrule
    Low parent ESP$_{\max}$ & +0.81 & +1.59 & +1.32 & +0.51 & +0.24 & +0.70 & +0.47 & +0.06 \\
    High parent ESP$_{\max}$ & +0.43 & +1.06 & +0.82 & +0.04 & -0.16 & +0.30 & +0.19 & -0.25 \\
    \bottomrule
  \end{tabular}
\end{table}

\FloatBarrier
\subsection{Depth-Resolved Distributions of Functional-Group Electronic Effects}
\label{si-sec:depth-resolved-distributions}

The main text characterizes each functional group by its median single-site response and by its median trajectory from one to four functionalized sites. Those summaries compress the behavior of thousands of molecules into a single value at each level; Figures~\ref{si-fig:si-parent-product-esp-n1234} and~\ref{si-fig:si-parent-product-hl-n1234} show the full distributions behind them, so that both the typical response and its spread can be read directly. Every functionalized molecule is compared with its corresponding parent scaffold, and each quantity is reported as a parent-referenced shift $\Delta X$ (Methods). In each figure the eight functional groups are arranged across the columns and the number of functionalized sites $n=1$--4 down the rows; filled points mark the median response and the nested contours enclose the 50\%, 70\%, and 90\% density regions. Per-panel sample sizes range from 1,652 responses (Me at $n=1$) to 6,532 (F at $n=4$) and partition the 161,081 paired-valid responses of the full library by functional-group class and number of functionalized sites.

\begin{figure}[htbp]
\centering
\includegraphics[width=\linewidth]{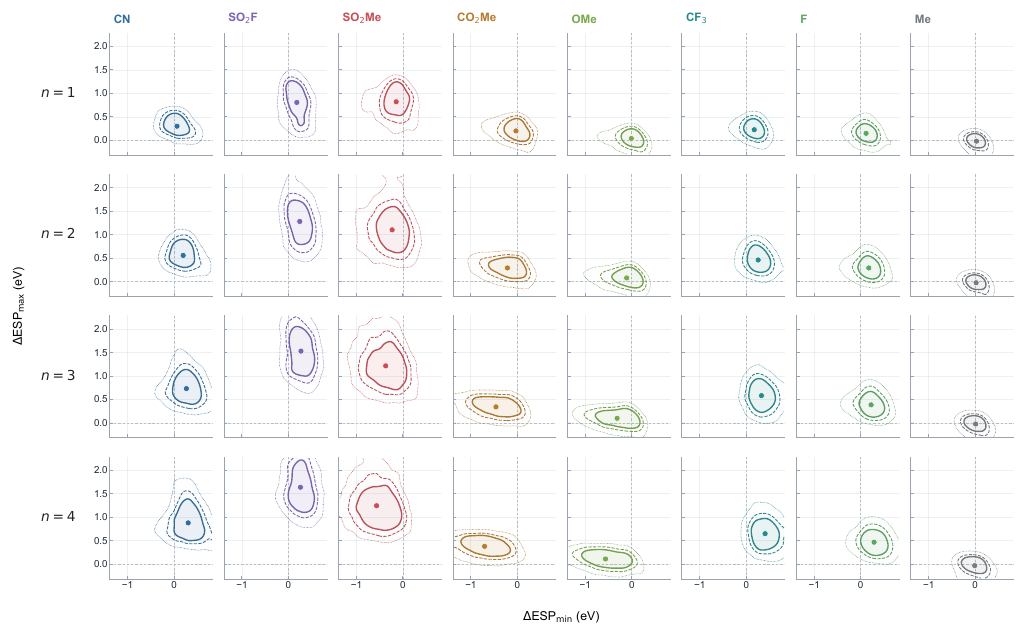}
\caption{\textbf{Parent-referenced electrostatic-potential shifts across one to four functionalized sites.} Distributions of $\Delta$ESP$_{\min}$ (horizontal) and $\Delta$ESP$_{\max}$ (vertical) for cyano (CN), fluorosulfonyl (SO$_2$F), methylsulfonyl (SO$_2$Me), methoxycarbonyl (CO$_2$Me), methoxy (OMe), trifluoromethyl (CF$_3$), fluoro (F), and methyl (Me) functionalization, one column per group. Rows correspond to $n=1$--4 functionalized sites (top to bottom), labeled at the left. Nested contours enclose the 50\%, 70\%, and 90\% density regions, filled points mark the median, and dashed lines mark zero shift from the parent.}
\label{si-fig:si-parent-product-esp-n1234}
\end{figure}

Electrostatic-potential extrema report how each functional group reshapes the local field of the scaffold (Figure~\ref{si-fig:si-parent-product-esp-n1234}). The sulfonyl groups impose the largest electron-poor excursions: SO$_2$F and SO$_2$Me already reach $\Delta$ESP$_{\max}\approx+0.81$ and $+0.82$ eV at $n=1$, and the SO$_2$F density core migrates steadily upward to a median of $+1.64$ eV by $n=4$, with CN following the same direction. Oxygen-bearing groups act in the opposite sense on the electron-rich extremum: OMe and CO$_2$Me deepen $\Delta$ESP$_{\min}$ progressively as functionalized sites accumulate (to $-0.56$ and $-0.70$ eV at $n=4$), marking increasingly pronounced Lewis-basic pockets, whereas F and CF$_3$ displace both extrema together along smooth, monotonic trajectories. Me stays clustered at the origin at every level and serves as the hydrocarbon reference.

\begin{figure}[htbp]
\centering
\includegraphics[width=\linewidth]{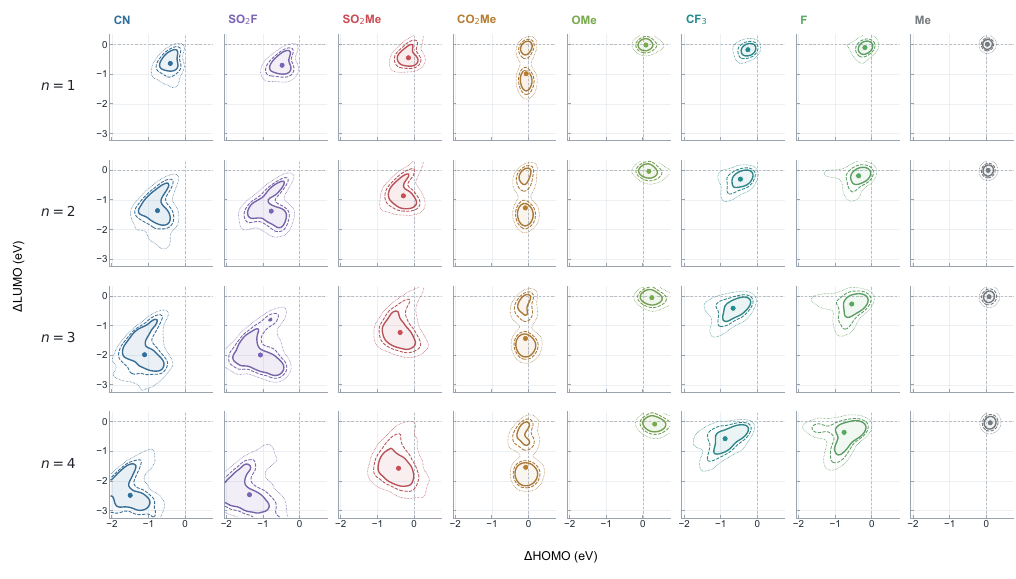}
\caption{\textbf{Parent-referenced frontier-orbital shifts across one to four functionalized sites.} Distributions of $\Delta$HOMO (horizontal) and $\Delta$LUMO (vertical) for the same eight functional groups and layout as Figure~\ref{si-fig:si-parent-product-esp-n1234}. Rows correspond to $n=1$--4 functionalized sites (top to bottom), labeled at the left; nested contours enclose the 50\%, 70\%, and 90\% density regions and filled points mark the median.}
\label{si-fig:si-parent-product-hl-n1234}
\end{figure}

The frontier-orbital distributions show distinct response patterns among the acceptor groups (Figure~\ref{si-fig:si-parent-product-hl-n1234}). CN and SO$_2$F converge on almost the same strongly stabilized LUMO at $n=4$ (median $\Delta$LUMO $-2.48$ and $-2.46$ eV) but follow different trajectories: CN also deepens the HOMO substantially ($-1.50$ eV) and increases the local positive field, whereas SO$_2$F couples its LUMO stabilization to the largest electrostatic shifts. CO$_2$Me primarily lowers the LUMO through the carbonyl $\pi^\ast$ orbital (median $\Delta$LUMO $-0.98$ eV at $n=1$ and $-1.54$ eV at $n=4$) while leaving the HOMO close to the parent value ($-0.08$ eV at $n=4$); its bimodal $\Delta$LUMO distribution is consistent with scaffold-dependent participation of the added ester group in the frontier system. OMe shows the complementary donor-side response, raising the HOMO (to $+0.32$ eV at $n=4$) with little LUMO change, while CF$_3$ and F accumulate moderate, monotonic stabilization of both levels (CF$_3$ $\Delta$HOMO from $-0.26$ to $-0.88$ eV) and Me remains near the reference. Across $n=1$--4, the density envelopes capture the progressive median shifts together with the scaffold-dependent spread.

\FloatBarrier
\subsection{Scaffold-Matched Comparison of Me, F, and \texorpdfstring{CF$_3$}{CF3} Functionalization}
\label{si-sec:cf3-f-me-comparison}

Single-site ($n=1$) methyl (Me), fluoro (F), and trifluoromethyl (CF$_3$) functionalization are compared through parent-referenced responses $\Delta X$ (Methods). The comparison draws on the same descriptor library as the preceding subsections, giving 6,785 CF$_3$/F/Me responses over 1,329 parent scaffolds (1,652 Me, 2,613 F, and 2,520 CF$_3$ responses). This scaffold-matched view describes the electronic changes associated with the first functionalized site.

\begin{figure}[htbp]
\centering
\includegraphics[width=\linewidth]{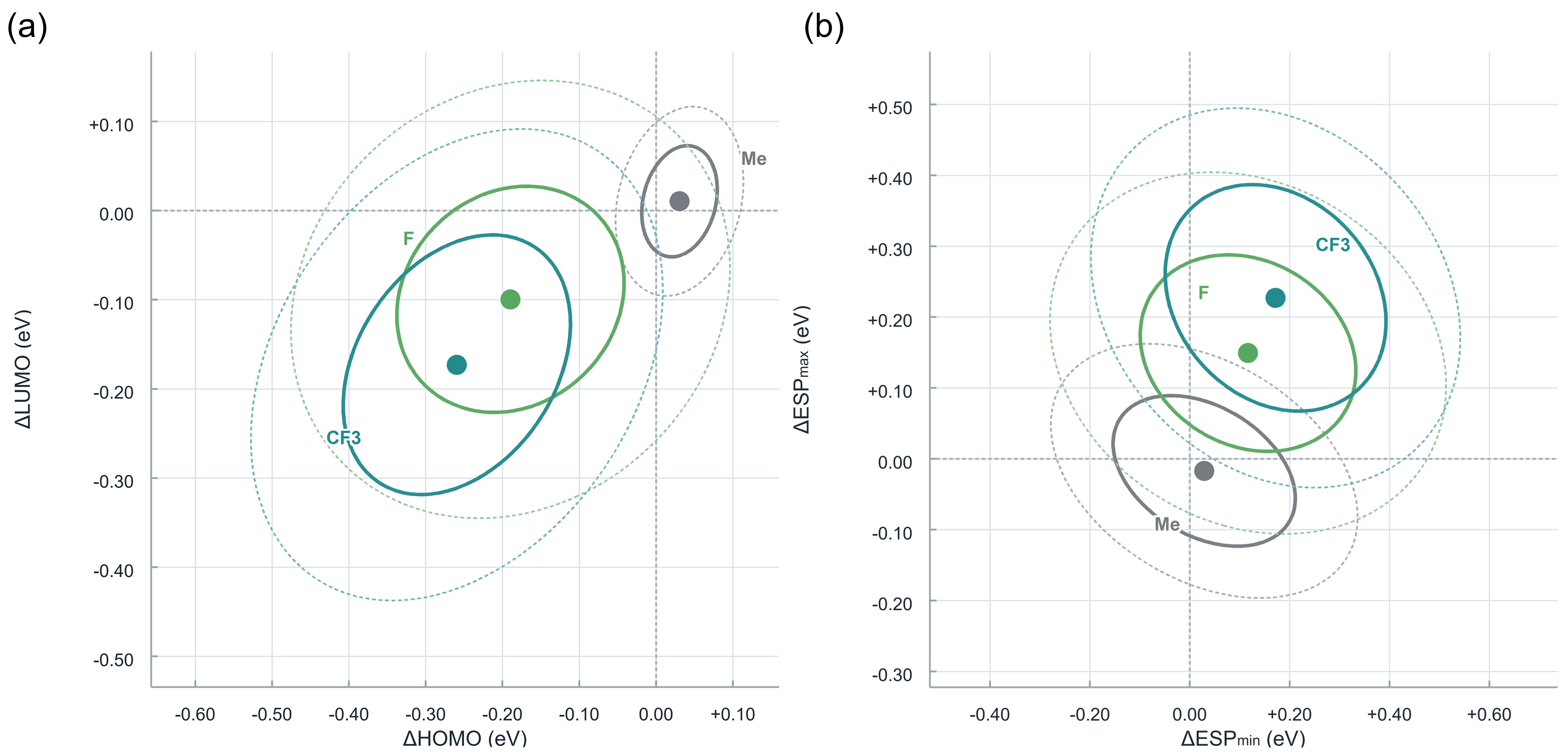}
\caption{\textbf{Scaffold-matched response planes for single-site Me, F, and CF$_3$ functionalization.} (a) Frontier-orbital plane: distributions of the parent-referenced shifts $\Delta$HOMO (horizontal) and $\Delta$LUMO (vertical) for the $n=1$ Me, F, and CF$_3$ subsets. (b) Electrostatic-potential plane: distributions of $\Delta$ESP$_{\min}$ (horizontal) and $\Delta$ESP$_{\max}$ (vertical) for the same subsets. In both panels, filled points mark the group medians, solid contours denote the 50\% core, and dashed contours the 80\% context region of each group.}
\label{si-fig:si-cf3-f-me}
\end{figure}

In the frontier-orbital plane (Figure~\ref{si-fig:si-cf3-f-me}a), Me remains close to the origin, with median $\Delta$HOMO/$\Delta$LUMO values of $+0.031$/$+0.010$ eV. F shifts the distribution center to $-0.190$/$-0.100$ eV, and CF$_3$ shifts it further to $-0.259$/$-0.173$ eV, consistent with the single-site medians in Table~\ref{si-tab:mono}. The ordering Me~$\rightarrow$~F~$\rightarrow$~CF$_3$ follows increasing fluorinated inductive withdrawal: fluorination lowers both frontier levels relative to the corresponding parent, with CF$_3$ giving the largest displacement and F occupying an intermediate regime. The density envelopes also show substantial scaffold-conditioned dispersion: the 50\% cores separate the median positions, whereas the 80\% envelopes retain overlap, especially between F and CF$_3$. Together, Me, F, and CF$_3$ provide a graded scaffold-matched comparison of frontier-level modulation.

The electrostatic-potential plane (Figure~\ref{si-fig:si-cf3-f-me}b) gives the complementary local-polarity view. Me is again near the origin, with median $\Delta$ESP$_{\min}$/$\Delta$ESP$_{\max}$ values of $+0.029$/$-0.017$ eV; F moves to $+0.117$/$+0.149$ eV, and CF$_3$ reaches $+0.172$/$+0.227$ eV. The simultaneous positive displacement of both extrema shows that fluorinated functionalization modifies the local electrostatic field together with the frontier-orbital levels. Across the two planes, Me anchors the hydrocarbon reference, F introduces moderate orbital lowering and electrostatic shifts, and CF$_3$ gives the larger fluorinated response. This scaffold-matched series compares frontier-level and local-field modulation in small-molecule electrolyte candidates.

\FloatBarrier
\section{Supplementary Statistics for \Li{} First-Shell Clusters}
\label{si-sec:clusters}

The cluster analysis covers all 22,500 relaxed structures. Every salt contributes 5,625 structures, every motif contributes 7,500 structures, and every salt--motif cell contributes 1,875 structures. All entries contain numerical HOMO, LUMO, gap, ESP extrema, total \Li{}--donor BO, \Li{} ELF deformation factor, pre-projection PENE, projected electron-number error, and fragment-resolved frontier-orbital weights. The reported statistics therefore summarize the complete 22,500-cluster library. Tables~\ref{si-tab:salt-summary}--\ref{si-tab:host-salt-group} collect the salt-pooled, motif-resolved, and functional-group-resolved statistics.

\begin{table}[htbp]
  \centering
  \footnotesize
  \setlength{\tabcolsep}{3pt}
  \caption{Salt-pooled descriptors for the 22,500-cluster library, 5,625 structures per salt.
  HOMO/LUMO solvent values are binary percentages under the conditional solvent--anion hosting
  definition given in Methods; BO entries are medians of the cumulative $\BOtot$ defined there and
  $r$ is the Pearson correlation between total BO and $\phi$. Orbital energies, the gap, and the
  ESP span (ESP$_{\max}-$ESP$_{\min}$) are medians in eV.}
  \label{si-tab:salt-summary}
  \begin{tabular}{@{}lrrrrrrrrr@{}}
    \toprule
    Salt & HOMO solv.\ (\%) & LUMO solv.\ (\%) & Median $\BOtot$ & Median $\phi$
    & $r(\BOtot,\phi)$ & HOMO & LUMO & Gap & ESP span \\
    \midrule
    LiPF$_6$ & 94.13 & 99.93 & 0.8113 & 1.03234 & -0.187 & -7.201 & -1.027 & 5.754 & 4.614 \\
    LiFSI    & 46.61 & 99.88 & 0.8627 & 1.03134 & +0.029 & -7.162 & -1.043 & 5.590 & 3.186 \\
    LiDFOB   & 28.07 & 71.08 & 0.8585 & 1.03058 & +0.015 & -7.175 & -1.488 & 5.358 & 3.767 \\
    LiTDI    & 0.00  & 91.61 & 0.6979 & 1.03199 & -0.082 & -5.476 & -1.756 & 3.156 & 6.462 \\
    \bottomrule
  \end{tabular}
\end{table}

\begin{table}[htbp]
  \centering
  \footnotesize
  \setlength{\tabcolsep}{3.5pt}
  \caption{Motif-resolved cluster statistics. Each row contains 1,875 structures. Orbital
  percentages refer to the solvent side of the conditional solvent--anion balance.}
  \label{si-tab:motif}
  \begin{tabular}{llrrrr}
    \toprule
    Salt & Motif & Median $\BOtot$ & Median $\phi$ & HOMO solvent (\%) & LUMO solvent (\%) \\
    \midrule
    LiPF$_6$ & M$_2$X & 0.8129 & 1.03178 & 94.51 & 100.00 \\
              & MBX & 0.8057 & 1.03193 & 92.37 & 100.00 \\
              & MX$_2$ & 0.8153 & 1.03351 & 95.52 & 99.79 \\
    \addlinespace
    LiFSI & M$_2$X & 0.8171 & 1.03187 & 50.03 & 100.00 \\
           & MBX & 0.8879 & 1.03079 & 49.23 & 100.00 \\
           & MX$_2$ & 0.8676 & 1.03133 & 40.59 & 99.63 \\
    \addlinespace
    LiDFOB & M$_2$X & 0.8525 & 1.03121 & 42.88 & 72.27 \\
            & MBX & 0.8479 & 1.03053 & 28.96 & 72.11 \\
            & MX$_2$ & 0.9484 & 1.02995 & 12.37 & 68.85 \\
    \addlinespace
    LiTDI & M$_2$X & 0.7010 & 1.03228 & 0.00 & 89.07 \\
           & MBX & 0.6964 & 1.03230 & 0.00 & 99.20 \\
           & MX$_2$ & 0.6980 & 1.03135 & 0.00 & 86.56 \\
    \bottomrule
  \end{tabular}
\end{table}

\begin{table}[htbp]
  \centering
  \small
  \caption{Pooled cluster medians by functional group across all four salts and three motifs. Gap and ESP span are in eV.}
  \label{si-tab:fg-cluster}
  \begin{tabular}{lrrrrr}
    \toprule
    Group & $N$ & Median $\BOtot$ & Median $\phi$ & Median gap & Median ESP span \\
    \midrule
    Me       & 1800 & 0.8256 & 1.03167 & 6.221 & 3.934 \\
    OMe      & 1800 & 0.8248 & 1.03114 & 6.199 & 3.801 \\
    F        & 2700 & 0.8377 & 1.03200 & 5.276 & 4.206 \\
    CF$_3$   & 3600 & 0.8364 & 1.03204 & 5.336 & 4.179 \\
    CO$_2$Me & 3600 & 0.8343 & 1.03174 & 5.335 & 4.047 \\
    CN       & 3600 & 0.8292 & 1.03167 & 5.123 & 4.159 \\
    SO$_2$F  & 2700 & 0.8064 & 1.03134 & 5.292 & 4.384 \\
    SO$_2$Me & 2700 & 0.8238 & 1.03105 & 5.101 & 4.000 \\
    \bottomrule
  \end{tabular}
\end{table}

Across the 22,500-structure library, the HOMO cation-dominant count is 0. For the LUMO, 550 structures (2.44\%) are cation-dominant, and 416 structures have cation weight above 50\%. These structures are included in the scalar and distributional summaries. The three-way count places cation contribution alongside the primary solvent-versus-anion coordinate used in the main text.

Table~\ref{si-tab:host-salt-group} resolves the hosting fractions of Table~\ref{si-tab:salt-summary} by functional-group class, and is the source of the salt-by-group values quoted in the main text. Reading down a column shows how far functionalization can move the hosting balance within one salt, and reading across a row shows how much of that balance is fixed by the salt itself.

\begin{table}[htbp]
  \centering
  \small
  \setlength{\tabcolsep}{4pt}
  \caption{Solvent-hosted fraction of the frontier orbitals (\%) for each salt and functional-group class, under the conditional solvent--anion hosting definition given in Methods. The complement of each entry is the anion-hosted fraction. Structure counts per cell are $n=900$ for CN, CO$_2$Me, and CF$_3$; $n=675$ for SO$_2$F, SO$_2$Me, and F; and $n=450$ for OMe and Me, summing to 5,625 per salt.}
  \label{si-tab:host-salt-group}
  \begin{tabular}{llrrrrrrrr}
    \toprule
    Orbital & Salt & CN & SO$_2$F & SO$_2$Me & CO$_2$Me & OMe & CF$_3$ & F & Me \\
    \midrule
    HOMO & LiPF$_6$ & 92.0 & 98.7 & 99.3 & 98.1 & 97.1 & 91.2 & 86.4 & 90.4 \\
     & LiFSI & 20.9 & 56.4 & 59.6 & 64.2 & 71.3 & 34.9 & 33.9 & 46.4 \\
     & LiDFOB & 5.4 & 46.4 & 46.4 & 37.8 & 55.3 & 13.2 & 11.7 & 26.0 \\
     & LiTDI & 0.0 & 0.0 & 0.0 & 0.0 & 0.0 & 0.0 & 0.0 & 0.0 \\
    \midrule
    LUMO & LiPF$_6$ & 100.0 & 100.0 & 100.0 & 100.0 & 99.6 & 100.0 & 100.0 & 99.6 \\
     & LiFSI & 100.0 & 100.0 & 100.0 & 100.0 & 99.1 & 100.0 & 100.0 & 99.3 \\
     & LiDFOB & 97.9 & 62.1 & 83.6 & 77.4 & 0.4 & 90.4 & 91.4 & 0.9 \\
     & LiTDI & 100.0 & 99.9 & 99.9 & 99.6 & 47.1 & 100.0 & 100.0 & 49.3 \\
    \bottomrule
  \end{tabular}
\end{table}

\FloatBarrier
\phantomsection
\addcontentsline{toc}{section}{References}
\section*{References}
\renewcommand{\bibsection}{}
\begin{multicols}{2}
\footnotesize
\setlength{\bibsep}{2.5pt}

\end{multicols}

\endgroup
\end{document}